\documentclass[10pt,journal,compsoc]{IEEEtran}
\pdfoutput=1

\ifCLASSOPTIONcompsoc
  \usepackage[nocompress]{cite}
\else
  \usepackage{cite}
\fi

\usepackage{times}
\usepackage{epsfig}
\usepackage{graphicx}
\usepackage{amsmath}
\usepackage{amssymb}
\usepackage{microtype}
\usepackage{color}
\usepackage{array}
\usepackage{hhline}
\usepackage{multirow}
\usepackage{caption}
\usepackage{subcaption}
\usepackage{enumitem}
\usepackage{tikz}
\usepackage{amsthm}
\def\checkmark{\tikz\fill[scale=0.4](0,.35) -- (.25,0) -- (1,.7) -- (.25,.15) -- cycle;}

\newcommand{\argmin}{\operatornamewithlimits{argmin}}
\newtheorem{proposition}{Proposition}

\begin{document}
\bstctlcite{IEEEexample:BSTcontrol}

\title{Deep Supervision with Intermediate Concepts}
%
%
%
%
\author{Chi~Li,
        ~M. Zeeshan~Zia,
        ~Quoc-Huy~Tran,
        ~Xiang~Yu,
        ~Gregory D.~Hager,
        and~Manmohan~Chandraker 
       
}

%
%


\IEEEtitleabstractindextext{%
\begin{abstract}
Recent data-driven approaches to scene interpretation predominantly pose inference as an end-to-end black-box mapping, commonly performed by a Convolutional Neural Network (CNN). However, decades of work on perceptual organization in both human and machine vision suggest that there are often intermediate representations that are intrinsic to an inference task, and which provide essential structure to improve generalization. In this work, we explore an approach for injecting prior domain structure into neural network training by supervising hidden layers of a CNN with intermediate concepts that normally are not observed in practice. We formulate a probabilistic framework which formalizes these notions and predicts improved generalization via this deep supervision method. One advantage of this approach is that we are able to train only from synthetic CAD renderings of cluttered scenes, where concept values can be extracted, but apply the results to real images. Our implementation achieves the state-of-the-art performance of 2D/3D keypoint localization and image classification on real image benchmarks including KITTI, PASCAL VOC, PASCAL3D+, IKEA, and CIFAR100. We provide additional evidence that our approach outperforms alternative forms of supervision, such as multi-task networks. 

\end{abstract}

}
\maketitle

\IEEEdisplaynontitleabstractindextext

%
\IEEEpeerreviewmaketitle

\IEEEraisesectionheading{\section{Introduction}\label{sec:introduction}}
%
%
%
%

\IEEEPARstart{O}{ur} visual world is rich in structural regularity. 
Studies in perception show that the human visual system imposes structure to reason about stimuli\cite{Smith_1986}. 
Consequently, early work in computer vision studied perceptual organization as a fundamental precept for recognition and reconstruction \cite{Lowe_1985,Marr_1982}. 
However, algorithms designed on these principles relied on hand-crafted features (e.g. corners or edges) and hard-coded rules (e.g. junctions or parallelism) to hierarchically reason about abstract concepts such as shape \cite{Mohan_Nevatia_1989,Sarkar_Soundararajan_2000}. 
Such approaches suffered from limitations in the face of real-world complexities. 
In contrast, convolutional neural networks (CNNs), as end-to-end learning machines, ignore inherent perceptual structures encoded by task-related intermediate concepts and attempt to directly map from input to the label space.  

Abu-Mostafa~\cite{abu1995hints} proposes ``hints'' as a middle ground, where a task-related hint derived from prior domain knowledge regularizes the training of neural networks by either constraining the parameter space or generating more training data.
In this work, we revisit and extend this idea by exploring a specific type of hint, which we refer to as an ``intermediate concept'', that encodes a sub-goal to achieve the main task of interest. 
For instance, knowing object orientation is a prerequisite to correctly infer object part visibility which in turn constrains the 3D locations of semantic object parts.
We present a generic learning architecture where intermediate concepts sequentially supervise hidden layers of a deep neural network to learn a specific inference sequence for predicting a final task. 

\begin{figure}[t]
  \centering
    \includegraphics[width=0.97\linewidth]{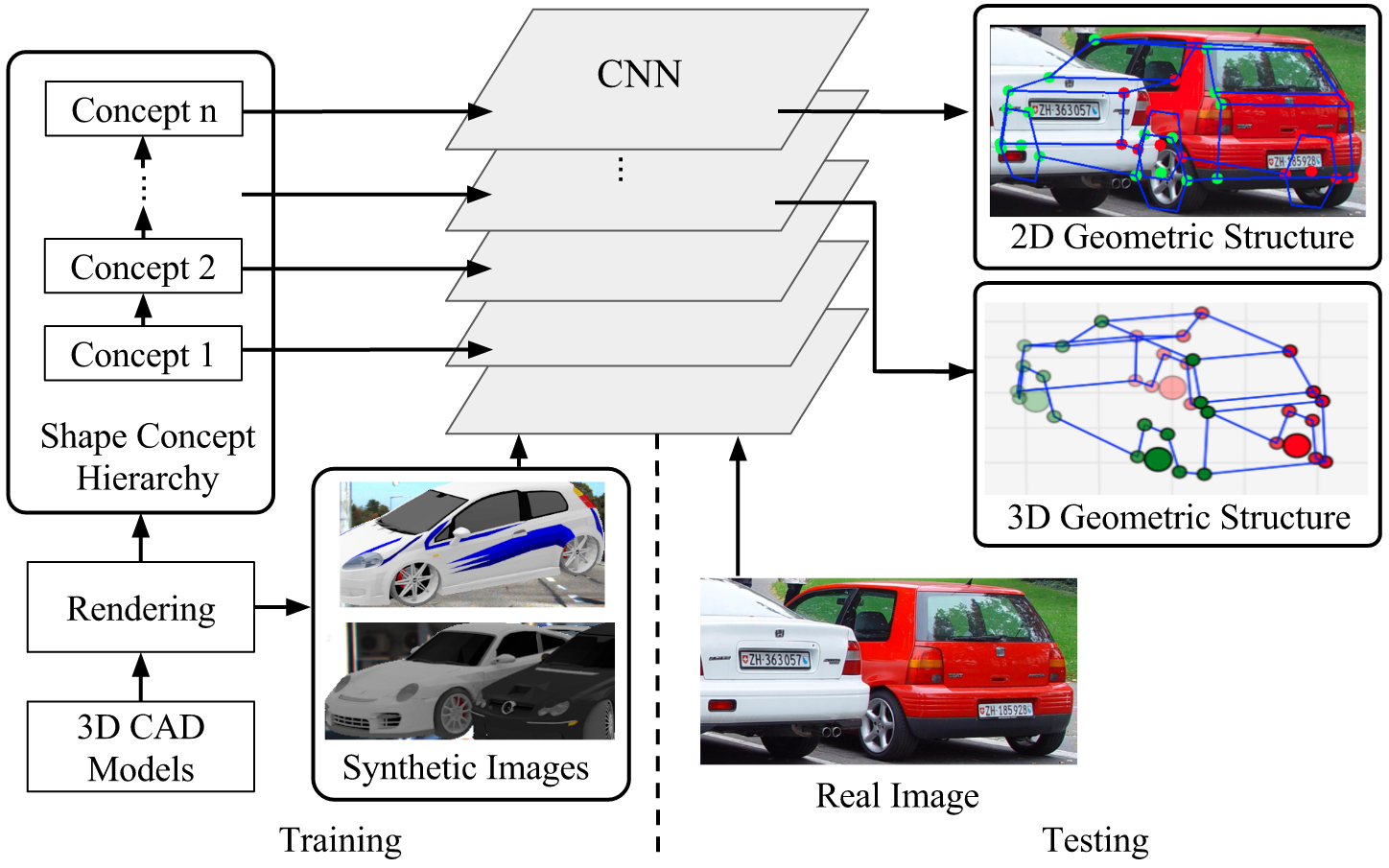}
  \caption{\small Overview of our approach. We use synthetic training images with intermediate shape concepts to deeply supervise the hidden layers of a CNN. At test time, given a single real image of an object, we demonstrate accurate localization of semantic parts in 2D and 3D, while being robust to intra-class appearance variations and occlusions.
  }
 \label{fig:intro}
\end{figure}

We implement this deep supervision framework with a novel CNN architecture for predicting 2D and 3D object skeletons given a single test image.
Our approach is in the spirit of \cite{Marr_1982,Lowe_1985} that exploit object pose as an auxiliary shape concept to aid shape interpretation and mental rotation.
We combine this early intuition with the discriminative power of modern CNNs by deeply supervising for multiple shape concepts such as object pose.
As such, deep supervision teaches the CNN to sequentially model intermediate goals to parse 2D or 3D object skeletons across large intra-class appearance variations and occlusion. 



An earlier version of this work has been presented in a conference paper~\cite{li2016deep}.
In this extended version, we formalize a probabilistic notion of intermediate concepts that predicts improved generalization performance by deeply supervising intermediate concepts (Section~\ref{sec:ds}). Further, we add new experiments including a new object class (bed) (Section~\ref{sec:ikea}) and image classification results on CIFAR100~\cite{krizhevsky09tr} (Section~\ref{sec:expcifar100}).
This motivates our network architecture in which we supervise convolutional layers at different depths with the available intermediate shape concepts.

Due to the scarcity of 3D annotated images, we render 3D CAD models to create synthetic images with concept labels as training data.
In addition, we simulate challenging occlusion configurations between objects to enable robust data-driven occlusion reasoning (in contrast to earlier model-driven attempts~\cite{xiang133dRR,zia2013detailed}).
Figure \ref{fig:intro} introduces our framework and Figure \ref{fig:pipe} illustrates an instance of a CNN deeply supervised by intermediate shape concepts for 2D/3D keypoint localization.
We denote our network as ``DISCO'' short for Deep supervision with Intermediate Shape COncepts.

Most existing approaches \cite{kar2015category,kulkarni2014inverse,wu2015learning,zhou2016learning,zia2013detailed} estimate 3D geometry by comparing projections of parameterized shape models with separately predicted 2D patterns, such as keypoint locations or heat maps. 
This makes prior methods sensitive to partial view ambiguity \cite{lee1985determination} and incorrect 2D structure prediction. 
Moreover, scarce 3D annotation of real image further limits their performance. 
In contrast, our method is trained on synthetic data only and generalizes well to real images. 
We find deep supervision with intermediate concepts to be a critical element to bridge the synthetic and real world.
In particular, our deep supervision scheme empirically outperforms the single-task architecture, and multi-task networks which supervise all the concepts at the final layer.
Further, we quantitatively demonstrate significant improvements over prior state-of-the-art for 2D/3D keypoint prediction on PASCAL VOC, PASCAL3D+\cite{xiang2014beyond}, IKEA\cite{lim2013parsing} 
and KITTI-3D where we add 3D annotation for part of KITTI\cite{geiger2012we} data.
These observations confirm that intermediate concepts regularize the learning of 3D shape in the absence of photorealism
in rendered training data.  

Additionally, we show another application of our generic deep supervision framework for image classification on CIFAR100~\cite{krizhevsky09tr}. 
As such, coarse-grained class labels used as intermediate concepts are able to improve fine-grained recognition performance, which further validates our deep supervision strategy.

In summary, we make the following contributions in this work:
\begin{itemize}
\item We present a CNN architecture where its hidden layers are supervised by a sequence of intermediate shape concepts for the main task of 2D and 3D object geometry estimation. 
\item We formulate a probabilistic framework to explain why deep supervision may be effective in certain cases. Our proposed framework is a generalization of conventional supervision schemes employed in CNNs, including multi-task supervision and Deeply Supervised Nets~\cite{lee2015deeply}.
\item We show the utility of rendered data with access to intermediate shape concepts. We model occlusions by rendering multiple object configurations, which presents a novel route to exploiting 3D CAD data for parsing cluttered scenes.
\item 
We empirically demonstrate state-of-the-art performance on 2D/3D semantic part localization and object classification on several public benchmarks. 
In some experiments, the proposed approach even outperforms the state-of-the-art methods trained on real images. We also demonstrate superior performance to baselines including the conventional multi-task supervision and different orders of intermediate concepts.
\end{itemize}

In the following, we review the related work in Section~\ref{sec:related} and introduce the probabilistic framework and algorithm of deep supervision in Section~\ref{sec:ds}.
Details of network architecture and data simulation are discussed in Section~\ref{sec:network_and_data}.
We discuss experiment results in Section~\ref{sec:exp} and conclude the paper in Section~\ref{sec:con}.

\section{Related Work}
\label{sec:related}
We present a deep supervision scheme with intermediate concepts for deep neural networks.
One application of our deep supervision is 3D object structure inference which is linked to recent advances including reconstruction, alignment and pose estimation. 
We review related work on these problems in the following:

\textbf{Multi-task Learning.} 
In neural networks, multi-task learning architectures exploit multiple task-related concepts to jointly supervise a network at the last layer. 
Caruana~\cite{caruana1998multitask} empirically demonstrates its advantage over a single-task neural architecture on various learning problems. 
Recently, multi-task learning has been applied to a number of vision tasks including face landmark detection~\cite{zhang2014facial} and viewpoint estimation~\cite{su2015render}. 
Hierarchy and Exclusion (HEX) graph~\cite{Deng2014eccv} is proposed to capture hierarchical relationships among object attributes for improved image classification.
In addition, some theories~\cite{baxter2000model,maurer2016benefit} attempt to investigate how shared hidden layers reduce 
required training data by jointly learning multiple tasks.
However, to our knowledge, no study has been conducted on quantifying the performance boost to a main task.
It is also unclear whether a design choice meets the assumption of conducive task relationships used in these theories.
This may explain that some task combinations for multi-task networks yield worse performance compared with single-task networks \cite{caruana1998multitask}.

\textbf{Deep Supervision.}
Deeply Supervised Nets (DSN) \cite{lee2015deeply} uses a single task label to supervise the hidden layers of a CNN, speeding up convergence and addressing the vanishing gradient problem.
However, DSN assumes that optimal local filters at shallow layers are building blocks for optimal global filters at deep layers, which is probably not true for a complex task.
Recently a two-level supervision is proposed~\cite{gulccehre2016knowledge} for counting objects in binary images.
One hidden layer is hard-coded to output object detection responses at fixed image locations.
This work can be seen as a preliminary study to leverage task-related cues that assist the final task by deep supervision.
We advance this idea further to a more general setting for deep learning without hard-coded internal representations. 


\textbf{3D Skeleton Estimation.} 
Many works model 3D shape as a linear combination of shape bases and optimize basis coefficients to fit computed image evidence such as heat maps \cite{zhou2016learning} and object part detections \cite{zia2013detailed}.
A prominent recent approach called single image 3D INterpreter Network (3D-INN)~\cite{wu2016single} is a sophisticated CNN architecture to estimate a 3D skeleton based only on detected visible 2D joints. 
However, in contrast to our approach, the training of 3D-INN does not jointly optimize for 2D and 3D keypoint localization. 
This decoupling of 3D structure from object appearance leads to partial view ambiguity and thus 3D prediction error.


\textbf{3D Reconstruction.}
A generative inverse graphics model is formulated in~\cite{kulkarni2014inverse} for 3D mesh reconstruction by matching mesh proposals to extracted 2D contours. 
Recently, given a single image, autoencoders have been exploited for 2D image rendering \cite{dosovitskiy2016learning}, multi-view mesh reconstruction \cite{tatarchenko2016multi} and 3D shape regression 
under occlusion \cite{pol2016overcoming}. The encoder network learns to invert the rendering process to recognize 3D attributes such as object pose. However, methods such as 
\cite{tatarchenko2016multi,pol2016overcoming} are quantitatively evaluated only on synthetic data and seem to achieve limited generalization to real images. Other works such 
as \cite{kar2015category} formulate an energy-based optimization framework involving appearance, keypoint and normal consistency for dense 3D mesh reconstruction, but require both 2D keypoint
 and object segmentation annotations on real images for training. Volumetric frameworks using either discriminative \cite{choy20163d} or generative \cite{rezende2016unsupervised} modeling infer a 3D shape distribution on voxel grids given image(s) of an object, limited to low-resolutions. 
Lastly, 3D voxel examplars \cite{xiang2015data} jointly recognize 3D shape and occlusion patterns by template matching, which is not scalable.    


\textbf{3D Model Retrieval and Alignment.}
This line of work estimates 3D object structure by retrieving the closest object CAD model and performing alignment, using 2D images 
\cite{aubry2014seeing,lim2014fpm,xiang2014beyond} and RGB-D data \cite{bansal2016marr,gupta2015inferring}. Unfortunately, 
a limited number of CAD models can not represent all instances in one object category.
Further, the retrieval step is slow for a large CAD dataset and alignment is sensitive to error in estimated pose.

\textbf{Pose Estimation and 2D Keypoint Detection.}
``Render for CNN''~\cite{su2015render} renders 3D CAD models as additional training data besides real images for object viewpoint estimation. 
We extend this rendering pipeline to support object keypoint prediction and cluttered scene rendering to learn occlusions from data. 
Viewpoint prediction is utilized in \cite{tulsiani2015viewpoints} to boost the  performance of 2D landmark localization. Recent work such 
as DDN~\cite{yu2016deep} optimizes deformation coefficients based on the PCA representation of 2D keypoints to achieve state-of-the-art performance on face 
and human body. 
Dense feature matching approaches which exploit top-down object category knowledge \cite{kanazawa2016warpnet,zhou2016learning} are recent successes, 
but our method yields superior results while being able to transfer knowledge from rich CAD data.



\textbf{Occlusion Modeling.}
Most work on occlusion invariant recognition relies on explicit occluder modeling~\cite{wang09iccv,zia2013detailed}. 
However, as it is hard to explicitly model object appearance, the variation in occluder appearance is also too broad to be captured effectively by model-driven approaches. 
This is why recent work has demonstrated gains by learning occlusions patterns from data~\cite{pepik13cvpr,xiang2015data}. 
Thanks to deep supervision, which enables effective generalization from CAD renderings to real images, we are able to leverage a significantly larger array of synthetic occlusion configurations.



\section{Deep Supervision with Intermediate Concepts}
\label{sec:ds}
\begin{figure}[t]
  \centering
    \includegraphics[width=0.8\linewidth]{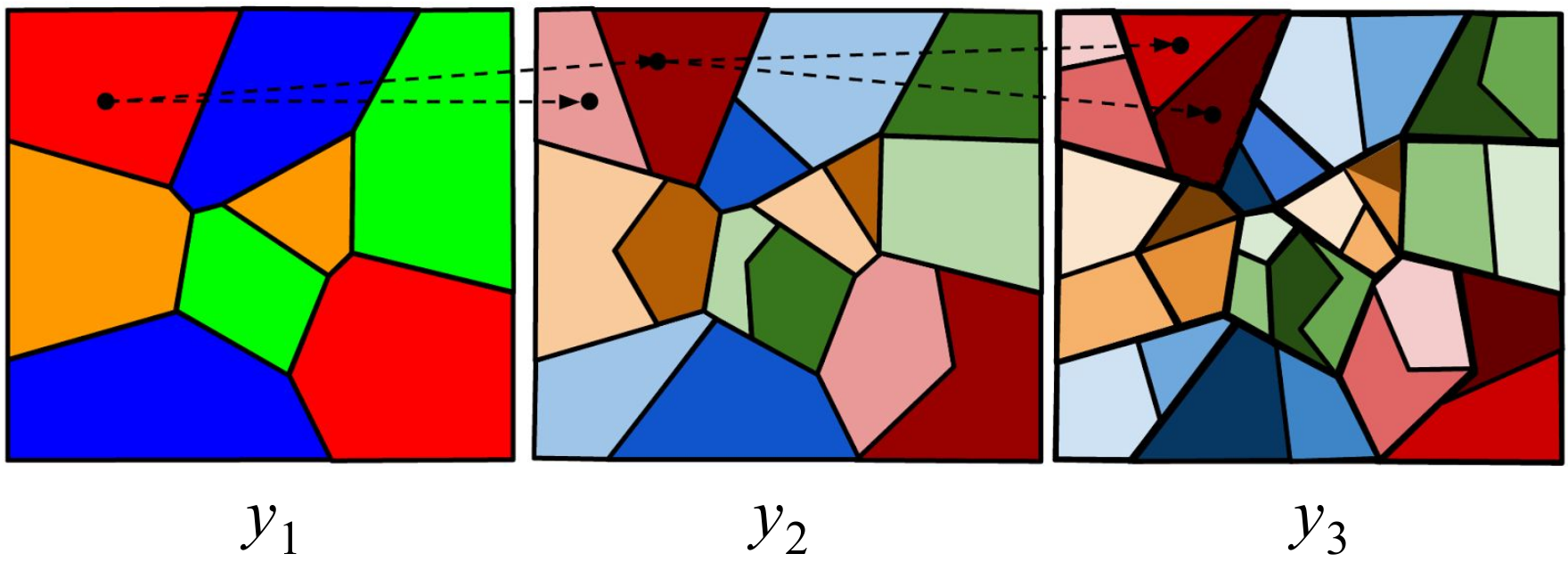}
  \caption{Illustration of a concept hierarchy with three concepts $\mathcal{Y}=\{y_1,y_2,y_3\}$ on 2D input space. Black arrows indicate the finer decomposition within the previous concept in the hierarchy. Each color represents one individual class defined by the concept.}
 \label{fig:ch}
 \vspace{-0.2cm}
\end{figure}

In this section, we introduce a novel CNN architecture with deep supervision.
Our approach draws inspiration from Deeply Supervised Nets (DSN) \cite{lee2015deeply}. 
DSN supervises each layer by the main task label to accelerate training convergence.
Our method differs from DSN in that we sequentially apply deep supervision on intermediate concepts intrinsic to the ultimate task, in order to regularize the network for better generalization.
We employ this enhanced generalization ability to transfer knowledge from richly annotated synthetic data to the domain of real images.

\textbf{Toy Example. }To motivate the idea of supervising intermediate concepts, consider a very simple network with 2 layers: $y = \sigma(w_2\sigma(w_1x+b_1)+b_2)$ where $\sigma$ is ReLU activation $\sigma(x)=\max(x,0)$.
Provided that the true model for a phenomenon is $(w_1,w_2,b_1,b_2)=(3,1,-2,-7)$ and the training data $\{(x,y)\}$ is $\{(1,0),(2,0),(3,0)\}$. 
A learning algorithm may obtain a different model $(w_1,w_2,b_1,b_2)=(1,3,-1,-10)$ which still achieves zero loss over training data but fails to generalize to the case when $x=4$ or $5$. 
However, if we have additional cues that tell us the value of intermediate layer activations, $\sigma(w_1x+b_1)$ for each $(x,y)$, we can achieve better generalization.
For example, suppose we have training examples with an additional intermediate cue $\{(x,y',y)\}=\{(1,0,0),(2,0,0),(3,1,0)\}$ where $y'=\sigma(w_1x+b_1)$.
We find that the incorrect solution above that works for $\{x,y\}$ is removed because it does not agree with $\{x,y',y\}$.
While simple, this example illustrates that deep supervision with intermediate concepts can regularize network training and reduce overfitting.

In the following, we formalize the notion of intermediate concept in Section \ref{sec:AC}, introduce our supervision approach which exploits intermediate concepts in Section \ref{sec:alg}, and discuss the improved generalization of deep supervision in Section \ref{sec:GA}.

\subsection{Intermediate Concepts}
\label{sec:AC}


We consider a supervised learning task to predict $y_m$ from $x$. 
We have a training set $S=\{(x,(y_1,\cdots,y_m))\}$ sampled from an unknown distribution $\mathcal{D}$, where
each training tuple consists of multiple task labels: $(y_1,\cdots,y_m)$. 
Without the loss of generality, we analyze the $i$-th concept $y_i$ in the following,  where $1<i\leq m$. Here, $y_{i-k}$ is regarded as an intermediate concept to estimate $y_i$, where $k>0$ and $i-k>0$.
Intuitively, knowledge of $y_{i-k}$ constrains the solution space of $y_i$, as in our simple example above. 


Formally, we define an intermediate concept $y_{i-k}$ of $y_i$ as a strict necessary condition such that there exists a deterministic function $T$ which maps $y_i$ to $y_{i-k}$: $y_{i-k}=T(y_i)$.
In general, there is no inverse function $T'$ that maps $y_{i-k}$ to $y_i$ because multiple $y_i$ may map to the same $y_{i-k}$.
In the context of multi-class classification where task $y_i$ and $y_{i-k}$ both contain discrete class labels, task $y_i$ induces a finer partition over the input space $\mathcal{X}=\{x\}$ than task $y_{i-k}$ by further partitioning each class in $y_{i-k}$.
Figure~\ref{fig:ch} illustrates a fictitious example of hierarchical partitioning over 2D input space created by three intermediate concepts $\{y_1,y_2,y_3\}$. 
As we can see in Figure~\ref{fig:ch}, a sequence of intermediate concepts hierarchically decompose the input space from coarse to fine granularity.
Concretely, we denote a concept hierarchy as $\mathcal{Y}=(y_1,\cdots,y_m)$ where $y_{i-k}$ is a strict necessary condition of $y_i$ for all $i>1$.

In many vision problems, we can find concepts that approximate a concept hierarchy $\mathcal{Y}$.
As mentioned above, non-overlapping coarse-grained class labels constitute strict necessary conditions for a fine-grained classification task. In addition, object pose and keypoint visibility are both strict necessary conditions for 3D object keypoint location, because the former can be unambiguously determined by the latter. 


\subsection{Algorithm}
\label{sec:alg}
\begin{figure}[t]
  \centering
    \includegraphics[width=0.9\linewidth]{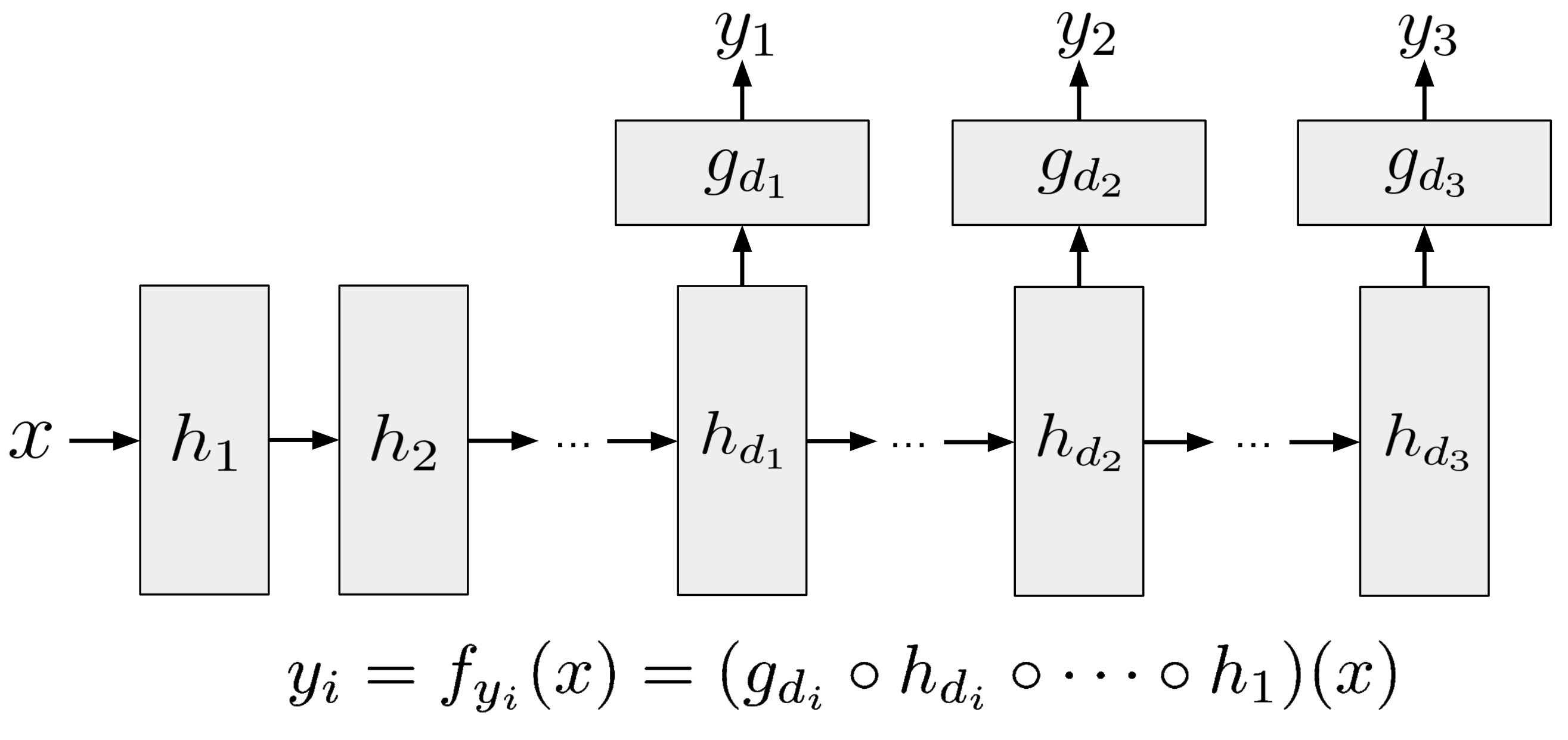}
  \caption{Schematic diagram of deep supervision framework. }
 \label{fig:ds_diagram}
  \vspace{-0.2cm}
\end{figure}

Given a concept hierarchy $\mathcal{Y}$ and the corresponding training set $S$, we formulate a new deeply supervised architecture to jointly learn the main task along with its intermediate concepts.
Consider a multi-layer convolutional neural network with $N$ hidden layers that receives input $x$ and outputs $m$ predictions for $y_1, \cdots, y_m$.
The $i$-th concept $y_i$ is applied to supervise the intermediate hidden layer at depth $d_i$ by adding a side output branch at $d_i$-th hidden layer.
We denote the function represented by the $k$-th hidden layer as $h_k(x, W_k)$, with parameters $W_k$.
The output branch at depth $d_i$ constructs a function $g_{d_i}(\cdot, V_{d_i})$ with parameters $V_{d_i}$. 
Further, we denote $f_{y_i}$ as the function for predicting concept $y_i$ such that $f_{y_i} = g_{d_i} \circ h_{d_i} \circ \cdots \circ h_1$.
Figure~\ref{fig:ds_diagram} shows a schematic diagram of our deep supervision framework.
In Section~\ref{sec:network_and_data}, we concretely instantiate each $h_k$ as a convolutional layer followed by batch normalization and ReLU layers and each $g_k$ as global average pooling followed by fully connected layers.
However, we emphasize that our algorithm is not limited to this particular layer configuration.


We formulate the following objective function to encapsulate these ideas:
\begin{equation}
W^*,V^*=\argmin_{W,V} \sum_{(x,\{y_i\})\in S} \sum_{i=1}^m \lambda_i l_i(y_i,f_{y_i}(x \ ; W_{1:d_i}, V_{d_i})) 
\label{eq:ds}
\end{equation}
\noindent
where $W_{1:d_i}=\{W_1, \cdots, W_{d_i}\}$, $W=W_{1:d_m}$ and $V=\{V_{d_1},\cdots,V_{d_m}\}$.  
In addition, $l_i$ is the loss for task $y_i$ scaled by the loss weight $\lambda_i$.
We optimize Equation~\ref{eq:ds} over $S$ by simultaneously backpropagating the loss of each supervisory signal all the way back to the first layer.

We note that Equation~\ref{eq:ds} is a generic supervision framework which represents many existing supervision schemes.
For example, the standard CNN with a single task supervision is a special case when $m=1$.
Additionally, the multi-task learning \cite{caruana1998multitask} places all supervision on the last hidden layer: $d_i=N$ for all $i$.
DSN\cite{lee2015deeply} framework is obtained when $m=N$ and $y_i=y_m$ for all $i$.
In this work, we propose to apply $m$ different concepts $\{y_i\}$ in a concept hierarchy $\mathcal{Y}$ at locations with growing depths: $d_{i-k} < d_i$ where $k>0$ and $i-k>0$.



\subsection{Generalization Analysis}
\label{sec:GA}

\begin{table}[t]
\small
\begin{tabular}{ |>{\centering\arraybackslash}m{0.4in} | >{\centering\arraybackslash}m{2.6in} |}
\hline
Notation 						& Meaning \\
\hline
$y_i$						 	& \multicolumn{1}{l|}{The $i$-th concept} \\[2pt]
$y_{i-k}$						& \multicolumn{1}{l|}{The intermediate concept of $y_i$} \\[2pt]
$d_i$ 						 	& \multicolumn{1}{l|}{The supervision depth of $y_i$} \\[2pt]
$f_{y_i}$						& \multicolumn{1}{l|}{A function that predicts $y_i$ given input $x$} \\[2pt]
$R(f_{y_i})$					& \multicolumn{1}{l|}{True risk of $f_{y_i}$} \\[2pt]
$R_S(f_{y_i})$					& \multicolumn{1}{l|}{Empirical risk of $f_{y_i}$ given a training set $S$}\\[2pt]
$\mathcal{H}_{y_i}$				& \multicolumn{1}{l|}{A set of $f_{y_i}$ with low empirical risk} \\[2pt]
$\mathcal{F}_{y_i}$				& \multicolumn{1}{l|}{A set of $f_{y_i}$ with low empirical and true risk} \\[2pt]
$\text{P}_{y_i}$				& \multicolumn{1}{l|}{Generalization probability of $y_i$} \\[2pt]
$\mathcal{H}_{y_i\mid y_{i-k}}$	& \multicolumn{1}{l|}{Subset of $\mathcal{H}_{y_i}$ that achieves low empirical risk on $y_{i-k}$} \\[2pt]
$\mathcal{F}_{y_i\mid y_{i-k}}$	& \multicolumn{1}{l|}{Subset of $\mathcal{F}_{y_i}$ that achieves low empirical risk on $y_{i-k}$} \\[2pt]
$\text{P}_{y_i\mid y_{i-k}}$	& \multicolumn{1}{l|}{Generalization probability $y_i$ constrained by $y_{i-k}$}	\\[2pt]
\hline
\end{tabular}
\caption{Notation table.}
\label{tab:notation}
\end{table}

In this section, we present a generalization metric and subsequently show how deep supervision with intermediate concepts can improve the generalization of a deep neural network with respect to this metric, compared to other standard supervision methods. 
We also discuss the limitations of this analysis.
For clarity, we summarize our notation in Table \ref{tab:notation}.

\subsubsection{Generalization Metric}
Deep neural networks are function approximators that learn mappings from an input space $x$ to an output space $y$.
For a network with a fixed structure, there usually exists a set of functions $\mathcal{H}$ (equivalently a set of parameters) where each element $f\in \mathcal{H}$ achieves a low empirical loss on a training set $S$. 
In the following, we define a generalization metric to measure the probability that a function $f\in \mathcal{H}$ is a ``true'' solution for a supervised learning task. 

Recall that $f_{y_i}$ represents the function composed by the first $d_i$ hidden layers and an output branch for predicting concept $y_i$.
The true risk $R(f_{y_i})$ is defined based on random variables $x$ and $y_i$ where $(x,y_i)\sim D$:
\begin{equation}
R(f_{y_i}) = \mathbb{E} \ [ l_i(f_{y_i}(x),y_i) \ ]
\label{eq:risk}
\end{equation}
Given a training set $S$, the empirical risk $R_S(f_{y_i})$ of $f_{y_i}$ is:
\begin{equation}
R_S(f_{y_i}) = \frac{1}{|S|}\sum_{(x,y_i)\in S} l_i(f_{y_i}(x),y_i)
\label{eq:emp}
\end{equation}

Given limited training data $S$, a deep neural network is optimized to find a solution $f_{y_i}$ with low empirical loss.
We consider empirical loss to be ``low'' when $R_S(f_{y_i}) < \delta$.
$\delta$ is the risk threshold which indicates ``good'' performance for a task.
Next, we define the function set $\mathcal{H}_{y_i}$ in which each function achieves low empirical risk:
\begin{equation}
\mathcal{H}_{y_i} = \{ f_{y_i} \mid R_S(f_{y_i}) < \delta \} 
\label{eq:H1}
\end{equation}
Similarly, we also define the function set $\mathcal{F}_{y_i}$ where each function achieves risks less than $\delta$ for both $R(f_{y_i})$ and $R_S(f_{y_i})$:
\begin{equation}
\mathcal{F}_{y_i} = \{ f_{y_i} \mid R_S(f_{y_i}) < \delta \ \wedge \ R(f_{y_i}) < \delta \} 
\label{eq:set}
\end{equation}
\noindent
By definition, we know $\mathcal{F}_{y_i}\subseteq \mathcal{H}_{y_i}$.
Given a training set and network structure, the generalization capability of the outcome of network training depends upon the likelihood that $f_{y_i}\in\mathcal{H}_{y_i}$ is also a member of $\mathcal{F}_{y_i}$. 

We consider $f_{y_i}$ to be a random variable as it is the outcome of a stochastic optimization process such as stochastic gradient descent. 
We assume that the optimization algorithm is unbiased within $\mathcal{H}_{y_i}$, such that apriori probability of converging to any $f_{y_i}\in\mathcal{H}_{y_i}$ is uniformly distributed.
We formalize a generalization metric for a CNN for predicting $y_i$ by defining a probability measure $\text{P}_{y_i}$ based on the function sets $\mathcal{F}_{y_i}$ and $\mathcal{H}_{y_i}$:
\begin{equation}
\begin{aligned}
\text{P}_{y_i} &= \text{P}(R(f_{y_i}) < \delta \mid R_S(f_{y_i}) < \delta) \\  
&=
\begin{cases}
        \frac{\mu(\mathcal{F}_{y_i})}{\mu(\mathcal{H}_{y_i})} & : \mathcal{H}_{y_i} \neq \emptyset \\
        0 & : \mathcal{H}_{y_i} = \emptyset \ 
\end{cases}
\end{aligned}
\label{eq:single_prob}
\end{equation}
\noindent
where $\mu(A)$ is the Lebesgue measure~\cite{lebesgue1902integrale} of set $A$ indicating the ``volume'' or ``size'' of set $A$ \footnote{Each function $f_{y_i}$ has a one-to-one mapping to a parameter $W$ in $\mathbb{R}^n$ where $n$ is the dimension of the parameter. We know that any subset of $\mathbb{R}^n$ is Lebesgue measurable.}. 
Moreover, $\mu(\mathcal{F}_{y_i})\leq \mu(\mathcal{H}_{y_i})$ due to $\mathcal{F}_{y_i}\subseteq \mathcal{H}_{y_i}$.
The equality $\mu(\mathcal{F}_{y_i}) = \mu(\mathcal{H}_{y_i})$ is achieved when $\mathcal{F}_{y_i}=\mathcal{H}_{y_i}$.
It follows that the higher the $\text{P}_{y_i}$, the better the generalization.

When an intermediate concept $y_{i-k}$ of $y_i$ is available, we insert one output branch $g_{d_{i-k}}$ at depth $d_{i-k}$ of CNN to predict $y_{i-k}$. 
Then, our deep supervision algorithm in Section~\ref{sec:alg} aims to minimize empirical risk on both $y_{i-k}$ and $y_i$.  
Recall that $f_{y_i}=g_{d_i}\circ f_{d_i}\circ \cdots \circ f_1$.
As a consequence, $f_{y_i}$ does not contain any output branch $g_{d_{i-k}}$ for the intermediate concept $y_{i-k}$. 
However, we note that $f_{y_i}$ shares some hidden layers with $f_{y_{i-k}}$. 
Similar to $\text{P}_{y_i}$, we can define the generalization probability $\text{P}_{y_i\mid y_{i-k}}$ of $f_{y_i}$ given the supervision of its intermediate concept $y_{i-k}$: 
\begin{equation}
\begin{aligned}
\text{P}_{y_i\mid y_{i-k}} &= \text{P}(R(f_{y_i})< \delta \mid R_S(f_{y_i})< \delta, R_S(f_{y_{i-k}})< \delta') \\
&= 
\begin{cases}
        \frac{\mu(\mathcal{F}_{y_i\mid y_{i-k}})}{\mu(\mathcal{H}_{y_i\mid y_{i-k}})} & : \mathcal{H}_{y_i\mid y_{i-k}} \neq \emptyset \\
        0 & : \mathcal{H}_{y_i\mid y_{i-k}} = \emptyset \ 
\end{cases}
\end{aligned}
\label{eq:prob}
\end{equation}
where the function set $\mathcal{H}_{y_i\mid y_{i-k}}$ is a subset of $\mathcal{H}_{y_i}$:
\begin{equation}
\mathcal{H}_{y_i\mid y_{i-k}} = \{f_{y_i}| R_S(f_{y_i}) < \delta \ \wedge \ R_S(f_{y_{i-k}}) < \delta'\} 
\label{eq:H2}
\end{equation}
and the function set $\mathcal{F}_{y_i\mid y_{i-k}}$ is a subset of $\mathcal{F}_{y_i}$:
\begin{equation}
\begin{aligned}
&\mathcal{F}_{y_i\mid y_{i-k}} = \{ f_{y_i}| R(f_{y_i}) < \delta \wedge R_S(f_{y_i}) < \delta \wedge R_S(f_{y_{i-k}}) < \delta' \} 
\label{eq:set}
\end{aligned}
\end{equation}
Note that we use a different threshold $\delta'$ for $R_S(f_{y_{i-k}})$ in order to account for the difference between loss functions $l_{i-k}$ and $l_i$.
We do not require the true risk of intermediate concept $R(y_{i-k})$ to be lower than $\delta'$ because the objective is to analyze the achievable generalization with respect to predicting $y_i$.

\subsubsection{Improved Generalization through Deep Supervision}
\label{sec:ig}
A machine learning model for predicting $y_i$ suffers from overfitting when the solution $f_{y_i}$ achieves low empirical risk $R_S(f_{y_i})$ over $S$ but high true risk $R(f_{y_i})$.
In other words, the higher the probability $\text{P}_{y_i}$, the lower the chance that the trained model $f_{y_i}$ overfits $S$. One general strategy to reduce the overfitting is to increase the diversity and size of training set $S$.
In this case, the denominator $\mu(\mathcal{H}_{y_i})$ of Equation~\eqref{eq:single_prob} decreases because fewer functions achieve low loss on more diverse data.
In the following, we show that supervising an intermediate concept $y_{i-k}$ of $y_i$ at some hidden layer is similarly capable of removing some incorrect solutions in $\mathcal{H}_{y_i}\setminus\mathcal{F}_{y_i}$ and thus improves the generalization because $\text{P}_{y_i\mid y_{i-k}}\geq\text{P}_{y_i}$.

First, given an intermediate concept $y_{i-k}$ of $y_i$ where $y_{i-k} = T(y_i)$, we specify the following assumptions for our analysis.
\begin{enumerate}
	\item \label{ass1} The neural network underlying our analysis is large enough to satisfy the universal approximation theorem \cite{hornik1989multilayer} for the concepts of interest, that is, its hidden layers have sufficient learning capacity to approximate arbitrary functions.
	\item \label{ass2} For a concept hierarchy $\mathcal{Y}=\{y_1, \cdots, y_m\}$, if $y'_i$ is a reasonable estimate of $y_i$, then $T(y'_i)$ should also be a reasonable estimate of the corresponding intermediate concept $y_{i-k}$.  Formally, we assume:
\begin{equation}
\forall y_i,y_i' \in Q_i: \ \ l_i(y_i,y'_i) \leq \delta \Rightarrow l_{i-k}(T(y_i),T(y'_i)) \leq \delta' 
\label{eq:loss_constraint}
\end{equation}
where $Q_i$ is the value space of concept $y_i$. 
	\item \label{ass3} Based on Assumption \ref{ass1} and \ref{ass2}, it follows that if $f_{y_i} \in \mathcal{F}_{y_i}$, there exists a $d_{i-k} < d_i$ such that the first $d_{i-k}$ layers of $f_{y_i}$ can be used to construct a $f_{y_{i-k}} \in \mathcal{F}_{y_{i-k}}$. 
\end{enumerate}
In practice, one may identify many tasks and relevant intermediate concepts satisfying Assumption \ref{ass2} when using common loss functions and $\delta=\delta'$. We discuss this further in Section \ref{sec:post}. 
To obtain Assumption \ref{ass3} above, we take the following two steps. 
First, with $k > 0$, Assumption \ref{ass1} allows us to find a $g_{d_{i-k}} = T \circ g_{d_i} \circ h_{d_i} \circ ~\cdots~ h_{d_{i-k+1}}$. As a consequence, we can always construct a $f_{y_{i-k}}$ from $f_{y_i}$ through $T$ using the first $d_{i-k}$ layers: $T \circ f_{y_i} = T \circ g_{d_i} \circ h_{d_i} \circ ~\cdots ~\circ h_1 = g_{d_{i-k}} \circ h_{d_{i-k}} \circ ~\cdots ~\circ h_1 = f_{y_{i-k}}$.
Second, Assumption \ref{ass2} further extends that for any $f_{y_i} \in \mathcal{F}_{y_i}$, its first $d_{i-k}$ layers can be used to obtain a $f_{y_{i-k}} \in \mathcal{F}_{y_{i-k}}$. 

Given an intermediate concept $y_{i-k}$ that satisfies the above assumptions, the following two propositions discuss how $d_{i-k}$ (the supervision depth of $y_{i-k}$) affects the generalization ability of $y_i$ in terms of $\text{P}_{y_i\mid y_{i-k}}$. 
First, we show that supervising intermediate concepts in the wrong order has no effect on improving  generalization.
\begin{proposition}
If $d_{i-k} \geq d_i$, the generalization performance of $y_i$ is not guaranteed to improve: 
\label{prop:core1}
\end{proposition}
\vspace{-0.15in}
\begin{equation}
\forall d_{i-k} \geq d_i, \ \ \ \text{P}_{y_i\mid y_{i-k}} = \text{P}_{y_i}
\end{equation}
\begin{proof}
We first consider the case when $y_i$ and $y_{i-k}$ both supervise the same hidden layer: $d_i=d_{i-k}$.
Given a sample set $(x,y_{i-k},y_i)\sim \mathcal{D}$ and a function $f_{y_i}$ which correctly predicts $y_i$ for $x$: $y_i=f_{y_i}(x)$, we can construct $f_{y_{i-k}}=T\circ f_{y_i}$ to yield the correct prediction for $y_{i-k}$. 
Based on Assumption \ref{ass1}, a multi-layer perceptron (i.e. fully connected layers) is able to represent any mapping function $T$.
Therefore, to approximate $f_{y_{i-k}}=T\circ f_{y_i}$, we can append fully connected layers which implement $T$ to $g_{d_i}$: $g_{d_{i-k}}=T\circ g_{d_i}$.
Based on Assumption \ref{ass2}, for any function $f_{y_i}$ in $\mathcal{F}_{y_i}$, there exists a corresponding function $f_{y_{i-k}}=T\circ f_{y_i}$ which satisfies $R_S(f_{y_{i-k}})\leq \delta'$.
This indicates that $\mathcal{H}_{y_i\mid y_{i-k}}=\mathcal{H}_{y_i}$ which in turn implies $\mathcal{F}_{y_i\mid y_{i-k}}=\mathcal{F}_{y_i}$.
When $d_{i-k}>d_i$, hidden layers from $d_i$ to $d_{i-k}$ can be implemented to achieve an identity mapping and then follow the same analysis for the case $d_i=d_{i-k}$. 
As a consequence, Proposition~\ref{prop:core1} holds.
\end{proof}

\begin{proposition}
There exists a $d_{i-k}$ such that $d_{i-k} < d_i$ and the generalization performance of $y_i$ is improved: 
\label{prop:core2}
\end{proposition}
\vspace{-0.15in}
\begin{equation}
\begin{aligned}
\exists d_{i-k} < d_i, \ \ \ \text{P}_{y_i\mid y_{i-k}} \geq \text{P}_{y_i}
\end{aligned}
\label{eq:core2}
\end{equation}
\begin{proof}
From Equation~\ref{eq:H1} and~\ref{eq:H2}, we observe that $\mathcal{H}_{y_i\mid y_{i-k}} \subset \mathcal{H}_{y_i}$ and $\mu(\mathcal{H}_{y_i\mid y_{i-k}}) < \mu(\mathcal{H}_{y_{i-k}})$.
Thus, we obtain:
\begin{equation}
\mu(\mathcal{H}_{y_i\mid y_{i-k}}) \leq \min(\mu(\mathcal{H}_{y_i}),\mu(\mathcal{H}_{y_{i-k}}))
\label{eq:set_reduce}
\end{equation}
Given a training set $S$, Equation~\ref{eq:set_reduce} essentially means that the number of functions that simultaneously fit both $y_i$ and $y_{i-k}$ is not more than the number of functions that fit each of them individually.
Intuitively, as the toy example earlier, the hidden layers of some network solutions for $y_i$ yield incorrect predictions of the intermediate concept $y_{i-k}$. 
This implies that $\mu(\mathcal{H}_{y_i\mid y_{i-k}}) \ll \min(\mu(\mathcal{H}_{y_i}),\mu(\mathcal{H}_{y_{i-k}}))$ in practice.
Subsequently, Assumption \ref{ass3} suggest that there exists one or multiple $d_{i-k}$'s such that the first $d_{i-k}$ layers of each solution $f_{y_i}\in \mathcal{F}_{y_i}$ are contained in $f_{y_{i-k}}\in \mathcal{F}_{y_{i-k}}$.
In other words, we can find a supervision depth $d_{i-k}$ for $y_{i-k}$ which satisfies:
\begin{equation}
\exists \ d_{i-k}<d_i, \ \ \ \mu(\mathcal{F}_{y_i}) = \mu(\mathcal{F}_{y_i|y_{i-k}}) 
\label{eq:uni}
\end{equation}
As a result, Proposition~\ref{prop:core2} is proved by Equation~\ref{eq:set_reduce} and Equation~\ref{eq:uni}. 
\end{proof}

To this end, we can improve the generalization of $y_i$ via $y_{i-k}$ by inserting the supervision of $y_{i-k}$ before $y_i$.
As a consequence, given a concept hierarchy $\mathcal{Y}_0=(y_1,\cdots,y_m)$, the supervision depths of concepts $\{d_1,\cdots,d_m\}$ should be monotonically increasing: $1\leq d_1<\cdots<d_m$. 
We then extend Equation~\ref{eq:set_reduce} to incorporate all available intermediate concepts of $y_m$:
\begin{equation}
\mu(\mathcal{H}_{y_m\mid y_{m-1},\cdots,y_1}) \leq \min_{y_i} \mu(\mathcal{H}_{y_i}) \ \ s.t. \ \ \forall i<j, d_i<d_j
\label{eq:core3}
\end{equation}
As we report in Section~\ref{sec:exp}, the empirical evidence shows that more intermediate concepts often greatly improves the generalization performance of the main task, which implies a large gap between two sides of Equation~\ref{eq:core3}.
Similar to Equation \ref{eq:uni}, we still have:
\begin{equation}
\exists \ d_1<\cdots<d_m, \ \ \ \mu(\mathcal{F}_{y_m}) = \mu(\mathcal{F}_{y_m|y_{m-1},\cdots,y_1}) 
\end{equation}
As a consequence, the generalization performance of $y_m$ given its necessary conditions $y_1,\cdots,y_{m-1}$ can be improved if we supervise each of them at appropriate depths $d_1,\cdots,d_{m-1}$ where $d_1<\cdots<d_{m-1}<d_m$:
\begin{equation}
\exists d_1 < \cdots < d_m, \ \ \ \text{P}_{y_m\mid y_{m-1},\cdots,y_1} \geq \text{P}_{y_m} 
\label{eq:core_final}
\end{equation}
Furthermore, $\text{P}_{y_m\mid y_{m-1},\cdots,y_1}$ is monotonically decreasing by removing intermediate concepts: $\text{P}_{y_m\mid y_{m-1},\cdots,y_1} \geq \text{P}_{y_m\mid y_{m-2},\cdots,y_1}\geq \cdots \geq \text{P}_{y_m\mid y_1} \geq \text{P}_{y_m}$.
The more concepts applied, the better chance that the generalization is improved. 
In conclusion, deep supervision with intermediate concepts regularizes the network training by decreasing the number of incorrect solutions that generalize poorly to the test set.

\subsubsection{Discussion}
\label{sec:post}

\begin{figure*}[t]
  \centering
    \includegraphics[width=0.95\linewidth]{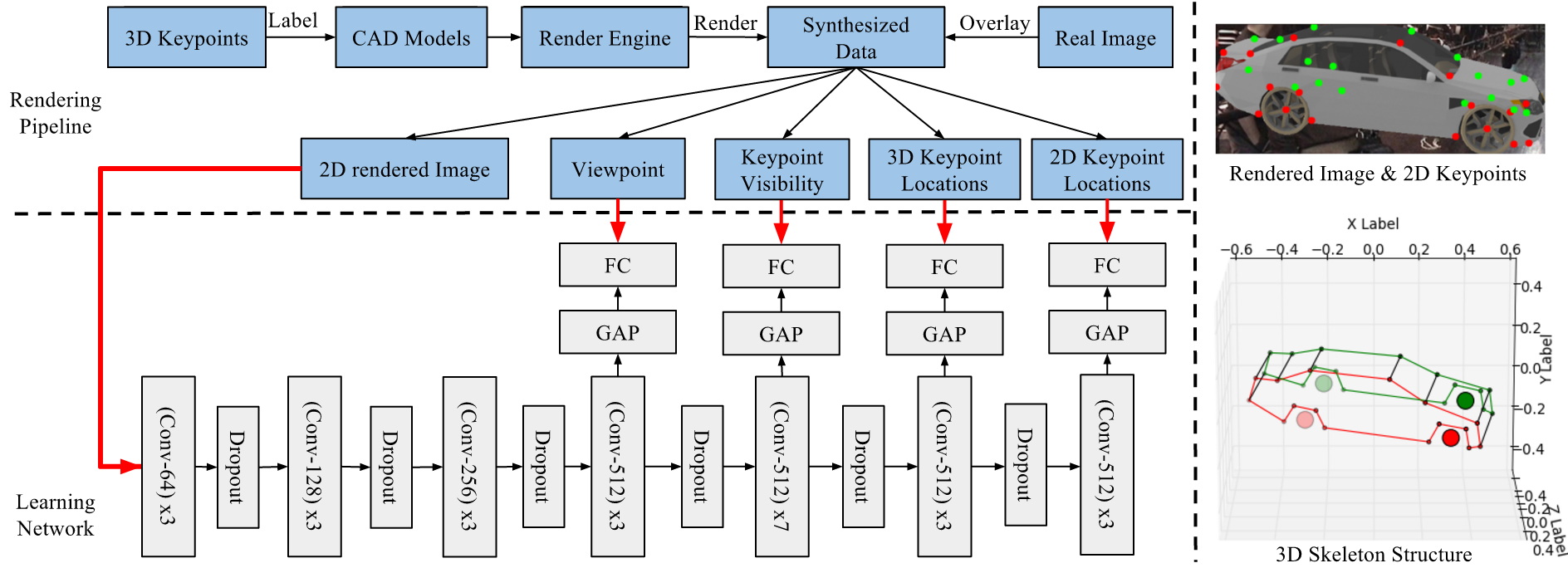}
  \caption{Visualization of our rendering pipeline (top-left), DISCO network (bottom-left), an example of rendered image and its annotations of 2D keypoints (top-right) as well as 3D skeleton (bottom-right).}
 \label{fig:pipe}
\end{figure*}

\textbf{Generalization of Intermediate Concept.}
We generalize the notion of intermediate concept, using conditional probabilities, with $y_{i-k}$ being the $\epsilon$-error necessary condition of $y_i$ if $y_{i-k}$ and $y_i$ for any sample $(x,(y_{i-k}, y_i)) \sim \mathcal{D}$ satisfy:
\begin{equation}
\forall c, \ \ \max \text{P}(y_{i-k} \mid y_{i}=c) \geq 1-\epsilon
\label{eq:ne1}
\end{equation}
\noindent
where $0 \leq \epsilon \leq 1$. 
The strict necessary condition defined in Section~\ref{sec:AC} holds when $\epsilon=0$.
When $\epsilon>0$, the monotonically increasing supervision order indicated by Equation~\ref{eq:core_final} is no longer ensured.
However, the architecture design suggested by our generalization analysis in Section~\ref{sec:ig} achieves the best performance in our empirical studies in Section~\ref{sec:exp}.
We believe that the generalization analysis in Section~\ref{sec:ig} is a good approximation for case with small $\epsilon$ in real applications.
We leave the analytic quantification of how $\epsilon$ affects deep supervision to future work.

\textbf{Assumption \ref{ass2}.}
If Assumption \ref{ass2} does not hold, both the numerator and denominator in Equation~\ref{eq:prob} decrease by different amounts. 
As a consequence, we cannot obtain Proposition~\ref{prop:core1} for all cases. 
However, many commonly used loss functions satisfy this assumption when $\delta=\delta'$.
One simple example is when $l_i$ and $l_{i-k}$ are indicator functions (i.e. $l_i(y,y')=\textbf{1}(y=y')$) for all $i$ \footnote{Note that the indicator function can be applied to discrete and continuous values of $y$ and $y'$.}.
As such, $l_i(y,y')=l_{i-k}(T(y),T(y'))$ when $\epsilon=0$ and thus Assumption \ref{ass2} is satisfied.
Another example can be that $l_i$ and $l_{i-k}$ are both L2 loss (i.e. $l_i(y,y')=\|y=y'\|^2$) and $T$ is a projection function where $T(y)=Py$ and $P$ is a projection (i.e. $P^2=P$).
In this case, $l_i(y,y') = \|y-y'\|^2 \geq \|P(y-y')\|^2 = l_{i-k}(T(y),T(y'))$. 

\textbf{Uniform Probability of $f_{y_i}\in \mathcal{H}_{y_i}$.}
In practice, this assumption may seem to contradict some empirical studies like~\cite{zhang2016understanding} where common CNNs generalize well after overfitting to large-scale training data (e.g. Imagenet~\cite{deng2009imagenet}).
This phenomenon actually demonstrates another dimension of improving generalization: training models on a large training set $S$ so that $\mathcal{H}_{y_i}$ is shrinking and converging to $\mathcal{F}_{y_i}$. 
Our work results shows that with deep supervision is an alternative route to achieve generalization given limited training data or data from a different domain, compared with standard supervision methods.

\textbf{DSN as a special case.}
Since a task is also a necessary condition of itself, our deep supervision framework actually contains DSN\cite{lee2015deeply} as a special case where each intermediate concept $y_i$ is the main task itself.
To illustrate the distinction enabled by our framework, we mimic DSN by setting the first intermediate concept $y_1=y_m$.
Thus, the first $d_1$ hidden layers are forced to directly predict $y_m$. 
Each $f_{d_1}\in \mathcal{F}_{y_1}$ can be trivially used to construct $f_{d_m}\in \mathcal{F}_{y_m}$ by forcing an identity function for layers $d_1$ to $d_m$.
This suggests that $\mathcal{F}_{y_m}$ is mainly constrained by $\mathcal{F}_{y_1}$.
Therefore, even though larger spatial supports from deeper layers between $d_1$ and $d_m$ reduce empirical risk in DSN, the learning capacity is restricted by supervision for $y_m$ at the first $d_1$ layers.


\section{Implementation and Data}
\label{sec:network_and_data}

We apply our method to both object classification and key point localization.  For object classification, we use the semantic hierarchy of labels to define intermediate concepts. 
For example, container is an intermediate concept (a generalization) of cup.  
For key point localization, 
we specify a 3D skeleton for each object class where nodes or keypoints represent semantic parts, and their connections define 3D object geometry.
Given a single real RGB image of an object, our goal is to predict the keypoint locations in image coordinates as well as normalized 3D coordinates while inferring their visibility states.
$X$ and $Y$ coordinates of 2D keypoint locations are normalized to $[0,1]$ along the image width and height, respectively.
3D keypoint coordinates are centered at origin and scaled to set the longest dimension along $X$,$Y$,$Z$ to unit length. 
Note that 2D/3D keypoint locations and their visibility all depend on the specific object pose with respect to the camera viewpoint.

To set up the concept hierarchy for 2D/3D keypoint localization, we have chosen in order, object orientation $y_1$, which is needed to predict keypoint visibility $y_2$, which roughly depicts the 3D structure prediction $y_3$, which finally leads to 2D keypoint locations $y_4$ including ones that are not visible in the current viewpoint.
We impose the supervision of the concept hierarchy $\mathcal{Y}=\{y_1,y_2,y_3,y_4\}$ into a CNN as shown in Fig.~\ref{fig:pipe} and minimize Equation~\ref{eq:ds} to compute the network parameters.

We emphasize that the above $\mathcal{Y}=\{y_1,y_2,y_3,y_4\}$ is not a $0$-error concept hierarchy because object pose ($y_1$), and 3D keypoint location ($y_3$) are not strict necessary conditions for visibility ($y_2$), and 2D keypoint location ($y_4$), respectively.
However, we posit that the corresponding residuals ($\epsilon$'s) of $\mathcal{Y}$ are small.
First, knowing object pose constrains keypoint visibilities to such an extent, that prior work has chosen to use ensembles of 2D templates for visual object parsing~\cite{li2011pami,pepik13cvpr}.
Second, there is a long and fruitful tradition in computer vision, starting from Marr's seminal ideas~\cite{Marr_1982} to leverage 3D object representations as a tool for 2D recognition.
In sum, our present choice of $\mathcal{Y}$ is an approximate realization of a $0$-error concept hierarchy which nonetheless draws inspiration from our analysis, and works well in practice.

\subsection{Network Architecture}
\label{sec:cnn}

In this section, we detail the network structure for keypoint localization. 
Our network resembles the VGG network \cite{simonyan2014very} and consists of deeply stacked $3\times 3$ convolutional layers.
Unlike VGG, we remove local spatial pooling between convolutional layers.
This is motivated by the intuition that spatial pooling leads to the loss of spatial information.
Further, we couple each convolutional layer with batch normalization \cite{ioffe2015batch} and ReLU, which defines $h_{d_i}(x,W_{d_i})$. 
The output layer $g_{d_i}(\cdot,V_{d_i})$ at depth $d_i$ for task $y_i$ is constructed with one global average pooling (GAP) layer followed by one fully connected (FC) layer with $512$ neurons, which is different from stacked FC layers in VGG.
The GAP layer averages filter responses over all spatial locations within the feature map.
From Table \ref{tab:car3d} in Section \ref{sec:kitti3d}, we empirically show that these two changes are critical to significantly improve the performance of VGG-like networks for 2D/3D landmark localization.

We follow the common practice of employing dropout \cite{krizhevsky2012imagenet} layers between the convolutional layers, as an additional means of regularization.
At layers $4$,$8$,$12$, we perform the downsampling using convolution layers with stride $2$.
The bottom-left of Figure \ref{fig:pipe} illustrates the details of our network architecture.
``(Conv-A)xB'' means A stacked convolutional layers with filters of size BxB.
We deploy $25$ convolutional layers in total.

We use L2 loss at all points of supervision. 
In practice, we only consider the azimuth angle of the object viewpoint with respect to a canonical pose. 
We further discretize the azimuth angle into $K$ bins and regress it to a one-hot encoding (the entry corresponding to the predicted discretized pose is set to $1$ and 
all others to $0$).  
Keypoint visibility is also represented by a binary vector with $1$ indicating occluded state of a keypoint.
During training, each loss is backpropagated to train the network jointly.

\subsection{Synthetic Data Generation}
\label{sec:data}

\begin{figure}[t]
  \centering
    \includegraphics[width=1.0\linewidth]{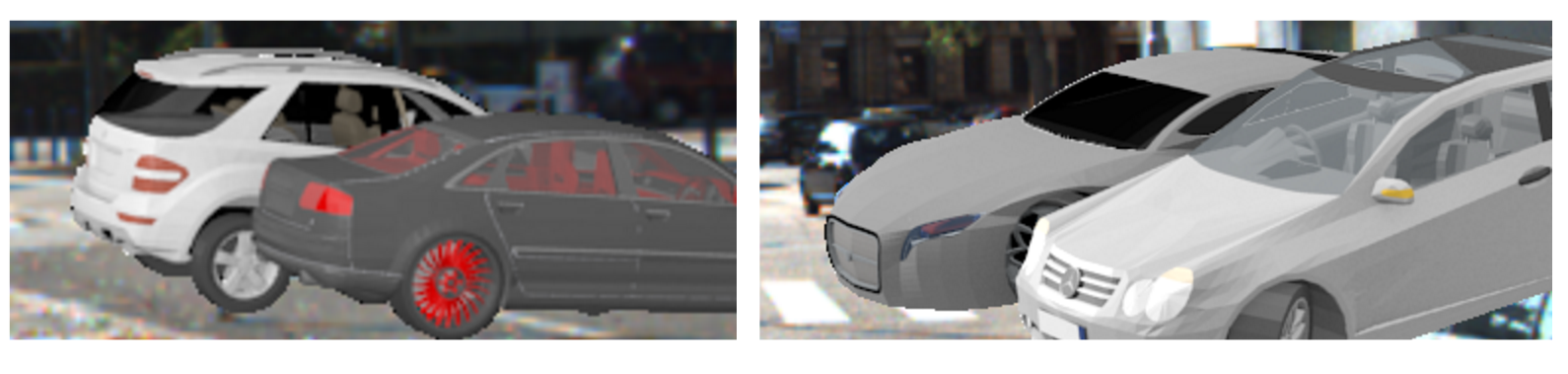}
  \caption{Examples of synthesized training images for simulating the multi-car occlusion.}
 \label{fig:occ_examples}
\end{figure}

Our approach needs a large amount of training data because it is based on deep CNNs. 
It also requests finer grained labels than many visual tasks such as object detection. 
Furthermore, we aim for the method to work for heavily cluttered scenes.
Therefore, we generate synthetic images that simulate realistic occlusion configurations involving multiple objects in close proximity. 
To our knowledge, rendering cluttered scenes that comprise of multiple CAD models is a novelty of our approach, although earlier work~\cite{pepik13cvpr,xiang2015data} used real image cut-outs for bounding box level localization.

An overview of the rendering process is shown in the upper-left of Fig.~\ref{fig:pipe}. 
We pick a small subset of CAD models from ShapeNet~\cite{chang2015shapenet} for a given object category and manually annotate 3D keypoints on each CAD model.
Next, we render each CAD model via Blender with randomly sampled graphics parameters including camera viewpoint, number/strength of light sources, and surface gloss reflection.
Finally, we follow \cite{su2015render} to overlay the rendered images on real backgrounds to avoid over-fitting.
We crop the object from each rendered image and extract the object viewpoint, 2D/3D keypoint locations and their visibility states from Blender as the training labels. 
In Figure \ref{fig:pipe} (right), we show an example of rendering and its 2D/3D annotations.

To model multi-object occlusion, we randomly select two different object instances and place them close to each other without overlapping in 3D space. 
During rendering, we compute the occlusion ratio of each instance by calculating the fraction of visible 2D area versus the complete 2D projection of CAD model. 
Keypoint visibility is computed by ray-tracing. 
We select instances with occlusion ratios ranging from $0.4$ to $0.9$. 
Fig.~\ref{fig:occ_examples} shows two training examples where cars are occluded by other nearby cars. 
For truncation, we randomly select two image boundaries (left, right, top, or bottom) of the object and shift them by $[0,0.3]$ of the image size along that dimension. \vspace{-1em}

\section{Experiments}
\label{sec:exp}

We first present an empirical study of image classification problem on CIFAR100~\cite{krizhevsky09tr} where a strict concept hierarchy is applied to boost the fine-grained object classification performance.
Subsequently, we extensively demonstrate competitive or superior performance for 2D/3D keypoint localization over several state-of-the-art methods, on multiple datasets: KITTI-3D, PASCAL VOC, PASCAL3D+ \cite{xiang2014beyond} and IKEA \cite{lim2013parsing}. 

\subsection{CIFAR100}
\label{sec:expcifar100}

\begin{table}[t]
\small
\centering
\begin{tabular}{| c | c |}
\hline
Methods & Error(\%) \\
\hline
DSN\cite{lee2015deeply} 		& 34.57 \\
FitNet, LSUV\cite{mishkin2015all} & 27.66 \\
ResNet-1001\cite{he2015deep} 	& 27.82 \\
pre-act ResNet-1001\cite{he2016identity} & 22.71 \\
\hline
plain-single					& 23.31 \\
plain-all						& 23.26 \\
DISCO-random					& 27.53 \\
DISCO							& \textbf{22.46} \\
\hline
\end{tabular}
\caption{Classification error of different methods on CIFAR100. The first four are previous methods and ``pre-act ResNet-1001'' is the current state-of-the-art. The remaining four are results of DISCO and its variants.
}
\label{tab:cifar100}
\end{table}

The image classification problem has a natural concept hierarchy where object categories can be progressively partitioned from coarse to fine granularity. 
In this section, we exploit coarse-grained class labels (20-classes) from CIFAR100~\cite{krizhevsky09tr} to assist fine-grained recognition into 100 classes.
Most existing methods directly learn a model for fine-grained classification task while ignoring coarse-grained labels.
In contrast, we leverage coarse-grained labels as an intermediate concept in our formulation.
We use the same network architecture shown in Section~\ref{sec:cnn} but with only 20 layers.
The number of filters are $128$, $256$ and $512$ for layers of 1-5, 6-10 and 10-20 respectively. 
Downsampling is performed at layer 6 and 11 and the coarse-grained label supervises layer 16.

Table \ref{tab:cifar100} compares the error of DISCO with state-of-the-art and variants of DISCO.
We use plain-single and plain-all to denote the networks with supervisions of single fine-grained label, and both labels at last layer, respectively.
DISCO-random uses a (fixed) random coarse-grained class label for each training image.
We observe that plain-all achieves roughly the same performance as plain-single, which replicates our earlier finding (Section~\ref{sec:kitti3d}) that intermediate supervision 
signal applied at the same layer as the main task helps relatively little in generalization.
However, DISCO is able to reduce the error of plain-single by roughly $0.6\%$ using the intermediate supervision signal. 
These results support our derivation of Proposition~\ref{prop:core1} and Proposition~\ref{prop:core2} in Section \ref{sec:GA}.
Further, DISCO-random is significantly inferior to DISCO as a random intermediate concept makes the training more difficult. 
Finally, DISCO slightly outperforms the current state-of-the-art ``pre-act ResNet-1001\cite{he2016identity}'' on image classification but with only half of the network parameters compared with \cite{he2016identity}.

\subsection{2D and 3D Keypoint Localization}

\begin{table*}[t]
\small
\centering
\begin{tabular}{|  >{\centering\arraybackslash}m{0.98in} |  >{\centering\arraybackslash}m{0.6in} |  >{\centering\arraybackslash}m{0.6in} | >{\centering\arraybackslash}m{0.9in} | >{\centering\arraybackslash}m{0.6in} | >{\centering\arraybackslash}m{0.6in} || >{\centering\arraybackslash}m{0.6in} | >{\centering\arraybackslash}m{0.6in} | >{\centering\arraybackslash}m{0.6in} |}
\hline
\multirow{2}{*}{Method}	&	\multicolumn{5}{c||}{2D} & 3D & 3D-yaw	\\
\hhline{|~|-|-|-|-|-|-|-|} & Full & Truncation & Multi-Car Occ & Other Occ & All & Full &Full \\
\hline
DDN~\cite{yu2016deep} & 67.6 & 27.2 & 40.7 & 45.0 & 45.1 & \multicolumn{2}{c|}{NA} \\
\hline
WN-gt-yaw*~\cite{kanazawa2016warpnet}  & 88.0 & 76.0 & 81.0 & 82.7 & 82.0 & \multicolumn{2}{c|}{NA} \\
\hline
Zia et al.~\cite{zia2013detailed} & 73.6 & \multicolumn{4}{c||}{NA} & 73.5 & 7.3 \\
\hhline{|=|=|=|=|=|=||=|=|}
DSN-2D     	  & 45.2 & 48.4 & 31.7 & 24.8 & 37.5 & \multicolumn{2}{c|}{NA} \\
\hline
DSN-3D     	  & \multicolumn{5}{c||}{NA} & 68.3 & 12.5 \\
\hline
plain-2D      & 88.4 & 62.6 & 72.4 & 71.3 & 73.7 & \multicolumn{2}{c|}{NA} \\
\hline
plain-3D      & \multicolumn{5}{c||}{NA} & 90.6 & 6.5 \\
\hline
plain-all     & 90.8 & 72.6 & 78.9 & 80.2 & 80.6 & 92.9 & 3.9 \\
\hline
DISCO-3D-2D	  & 90.1 & 71.3 & 79.4 & 82.0 & 80.7 & 94.3 & 3.1 \\ 
\hline
DISCO-vis-3D-2D & 92.3 & 75.7 & 81.0 & 83.4 & 83.4 & 95.2 & 2.3 \\
\hline
DISCO-(3D-vis) & 91.9 & 77.6 & 82.2 & \textbf{86.1} & 84.5 & 94.2 & 2.3 \\
\hline
DISCO-reverse & 30.4 & 29.7 & 22.8 & 19.6 & 25.6 & 54.8 & 13.0 \\
\hline
DISCO-Vgg     & 83.5 & 59.4 & 70.1 & 63.1 & 69.0 & 89.7 & 6.8 \\ 
\hline
DISCO	 	  & \textbf{93.1} & \textbf{78.5} & \textbf{82.9} & 85.3 & \textbf{85.0} & \textbf{95.3} & \textbf{2.2} \\ 
\hhline{|=|=|=|=|=|=||=|=|}
DISCO(Det)     & 95.9 & 78.9 & 87.7 & 90.5 & 88.3 & 95.5 & 2.1 \\ 
\hline
\end{tabular}
\caption{\small PCK[$\alpha=0.1$] accuracies (\%) of different methods for 2D and 3D keypoint localization on KITTI-3D dataset. Last column represents angular error in degrees. WN-gt-yaw~\cite{kanazawa2016warpnet} uses groundtruth pose of the test car. The bold numbers indicates the best result on groundtruth object bounding boxes. The last row presents the accuracies of DISCO on detection results from RCNN\cite{girshick2014rich}. }
\label{tab:car3d}
\end{table*}

\begin{table}[t]
\small
\centering
\begin{tabular}{ |>{\centering\arraybackslash}m{0.4in}  >{\centering\arraybackslash}m{0.4in}  >{\centering\arraybackslash}m{0.4in}|  >{\centering\arraybackslash}m{0.4in} | >{\centering\arraybackslash}m{0.4in} |  >{\centering\arraybackslash}m{0.4in} |}
\hline
\multicolumn{3}{|c|}{Training Data} & \multicolumn{3}{c|}{Test Data} \\
\hhline{|-|-|-|-|-|-|} 
\multicolumn{1}{|c|}{Full}  & \multicolumn{1}{c|}{Trunc.}  & \multicolumn{1}{c|}{Multi-Car} & Full & Trunc. & Occ. \\
\hline
\checkmark & &     & 91.8 & 53.6 & 68.3 \\
\hline
 & \checkmark &    & 89.9 & 73.8 & 61.7 \\
\hline
 & & \checkmark    & 91.3 & 74.7 & 82.7 \\
\hline
\checkmark & \checkmark &    & 92.9 & 71.3 & 63.4 \\
\hline
\checkmark & & \checkmark    & 92.5 & 73.2 & \textbf{84.1} \\
\hline
& \checkmark & \checkmark    & 90.5 & 70.4 & 81.2 \\
\hline
\checkmark & \checkmark & \checkmark & \textbf{93.1} & \textbf{78.5} & 83.2 \\
\hline
\end{tabular}
\caption{Ablative study of different training data sources. PCK[$\alpha=0.1$] accuracies (\%) of DISCO for 2D keypoint localization on KITTI-3D dataset.
}
\label{tab:data_ablation}
\end{table}

In this Section, we demonstrate the performance of the deep supervision network (Fig. \ref{fig:pipe}) for predicting the locations of object keypoints on 2D image and 3D space.

\textbf{Dataset.}
For data synthesis, we sample CAD models of $472$ cars, $100$ sofas, $100$ chairs and $62$ beds from ShapeNet \cite{chang2015shapenet}. 
Each car model is annotated with $36$ keypoints~\cite{zia2013detailed} and each furniture model (chair, sofa or bed) with $14$ keypoints~\cite{xiang2014beyond} \footnote{We use 10 keypoints which are consistent with \cite{wu2016single} to evaluate chair and bed on IKEA.}.
We synthesize 600k car images including occluded instances and 300k images of fully visible furniture (chair+sofa+bed).
We pick rendered images of 5 CAD models from each object category as validation set.

We introduce KITTI-3D with annotations of 3D keypoint and occlusion type on $2040$ car images from~\cite{geiger2012we}. 
We label car images with one of four occlusion types: no occlusion (or fully visible cars), truncation, multi-car occlusion (target car is occluded by other cars) and occlusion cause by other objects. 
The number of images for each type is $788$, $436$, $696$ and $120$, respectively. 

To obtain 3D groundtruth for these car images, we fit a PCA model trained on 3D keypoint annotation on CAD data, by minimizing the 2D projection error for known 2D landmarks provided by Zia et al.~\cite{zia2013detailed} and object pose from KITTI \cite{geiger2012we}. 
First, we compute the mean shape $M$ and $5$ principal components $P_1, \cdots, P_5$ from 3D skeletons of our annotated CAD models. 
$M$ and $P_i$ ($1\leq i \leq 5$) are $3\times 36$ matrices where each column contains 3D coordinates of a keypoint.
Thus, the 3D object structure $X$ is represented as $X=M+\sum_{i=1}^5\alpha_i P_i$, where $\alpha_i$ is the weight for $P_i$. 
To avoid distorted shapes caused by large $\alpha_i$, we constrain $\alpha_i$ to lie within $-2.7\sigma_i\leq \alpha_i\leq 2.7\sigma_i$ where $\sigma_i$ is the standard deviation along the $i^{th}$ principal component direction.
Next, given the groundtruth pose $T$, we compute 3D structure coefficients $\alpha = \{\alpha_i\}$ that minimize the projection error with respect to 2D ground truth $Y$:
\begin{equation}
\begin{aligned}
\alpha^* 
= &\arg_{\alpha} \min_{s,\beta} \| s\text{Pr}(T(M+\sum_{i=1}^N\alpha_i P_i))  + \beta - Y \|_2^2 \\ 
&s.t. \ -2.7\sigma_i\leq \alpha_i\leq 2.7\sigma_i  
\label{eq:gt3d}
\end{aligned}
\end{equation}
where the camera intrinsic matrix is $K=[s_x,0,\beta_x;0,s_y,\beta_y;0,0,1]$ with the scaling $s=[s_x;s_y]$ and shifting $\beta=[\beta_x;\beta_y]$.
$\text{Pr}(x)$ computes the 2D image coordinate from 2D homogeneous coordinate $x$.
In practice, to obtain the ground truth with even higher quality, we densely sample object poses $\{T_j\}$ in the neighborhood of $T$ and solve \eqref{eq:gt3d} by optimizing $\{\alpha_i\},\beta,s$ given a fixed $T_j$ and then search for the lowest error among all sampled $T_j$.
We only provide 3D keypoint labels for fully visible cars because we do not have enough visible 2D keypoints for most of the occluded or truncated cars and thus obtain rather crude 3D estimates for such cases.

\textbf{Evaluation metric.}
We use PCK and APK metrics~\cite{yang2011articulated} to evaluate the performance of 2D keypoint localization. 
A 2D keypoint prediction is correct when it lies within the radius $\alpha L$ of the ground truth, where $L$ is the maximum of image height and width and $0 < \alpha < 1$. 
PCK is the percentage of correct keypoint predictions given the object location and keypoint visibility.
APK is the mean average precision of keypoint detection computed by associating each estimated keypoint with a confidence score. 
In our experiments, we use the regressed values of keypoint visibility as confidence scores. 
We extend 2D PCK and APK metrics to 3D by defining a correct 3D keypoint prediction whose euclidean distance to the ground truth is less than $\alpha$ in normalized coordinates.

\textbf{Training details.}
We set loss weights of visibility, 3D and 2D keypoint locations $\{\lambda_i\}$ to $1$ and object pose to $0.1$.
We use stochastic gradient descent with momentum $0.9$ to train the proposed CNN from scratch. 
Our learning rate starts at $0.01$ and decreases by one-tenth when the validation error reaches a plateau. 
We set the weight decay to $0.0001$, resize all input images to $64$x$64$ and use batch size of $100$.
We initialize all weights using Glorot and Bengio~\cite{glorot2010understanding}.
For car model training, we form each batch using a mixture of fully visible, truncated and occluded cars, numbering $50$, $20$ and $30$, respectively. 
For the furniture, each batch consists of $70$ fully visible and $30$ truncated objects randomly sampled from the joint synthetic image set of chair, sofa and bed.

\subsubsection{KITTI-3D}
\label{sec:kitti3d}

We compare our method with DDN \cite{yu2016deep} and WarpNet \cite{kanazawa2016warpnet} for 2D keypoint localization and Zia et al.~\cite{zia2013detailed} for 3D structure prediction.
We use the original source codes for these methods.
However, WarpNet is a siamese archtecture which warps a reference image to a test image benefiting from class-aware training. In order to use it for landmark transfer task,
 we need a reference image to be warped. Thus, we retrieve $30$ labeled synthetic car images with the same pose as test image for landmark transfer using the CNN architecture proposed in~\cite{kanazawa2016warpnet} (WN-gt-yaw),
 and then compute the median of predicted landmark locations as the final result. The network is trained to warp pairs of synthetic car images in similar poses.
Additionally, we perform an ablative analysis of DISCO. 
First, we replace all intermediate supervisions with the final labels, as DSN~\cite{lee2015deeply} does, for 2D (DSN-2D) and 3D (DSN-3D) structure prediction. 
Next, we incrementally remove the deep supervision used in DISCO one by one. 
DISCO-vis-3D-2D, DISCO-3D-2D, plain-3D, and plain-2D are networks without pose, pose+visibility, pose+visibility+2D and pose+visibility+3D, respectively. 
Further, we change the locations of the intermediate supervision signals. 
plain-all shifts supervision signals to the final convolutional layer. 
DISCO-(3D-vis) switches 3D and visibility in DISCO, and DISCO-reverse reverses the entire order of supervisions in DISCO. 
Finally, DISCO-VGG replaces stride-based downsampling and GAP in DISCO with non-overlapping spatial pooling ($2$x$2$) and a fully connected layer with $512$ neurons, respectively.
All methods are trained on the same set of synthetic training images and tested on real cropped cars on ground truth locations in KITTI-3D.

In Table \ref{tab:car3d}, we report PCK accuracies for various methods\footnote{We cannot report Zia et al.\cite{zia2013detailed} on occluded data because only a subset of images has valid result in those classes.} and the mean error of estimated yaw angles ``3D-yaw'' over all fully visible cars.
This object-centric yaw angle is computed by projecting all 3D keypoints onto the ground plane and averaging the directions of lines connecting correspondences between left and right sides of a car.
In turn, the 3D-yaw error is the average of absolute error between the estimated yaw and the ground truth.

\begin{table*}[t]
\small
\centering
\begin{tabular}{ |>{\centering\arraybackslash}m{0.8in} | >{\centering\arraybackslash}m{0.6in} | >{\centering\arraybackslash}m{0.8in} | >{\centering\arraybackslash}m{0.6in} | >{\centering\arraybackslash}m{0.6in}  | >{\centering\arraybackslash}m{0.6in} | >{\centering\arraybackslash}m{0.6in}  | >{\centering\arraybackslash}m{0.9in} |}
\hline
PCK[$\alpha=0.1$] & Full & Full[$\alpha=0.2$] & Occluded & Big Image & Small Image & All [APK $\alpha=0.1$] \\
\hline
Long\cite{long2014convnets}  & 55.7 & \multicolumn{5}{c|}{NA} \\  
\hline
VpsKps\cite{tulsiani2015viewpoints}  & 81.3 & 88.3 & \textbf{62.8} & \textbf{90.0} & 67.4 & 40.3 \\
\hhline{|=|=|=|=|=|=|=|}
DSN-2D 			& 75.4 & 87.8 & 54.5 & 85.5 & 63.3 & NA \\
plain-2D 		& 76.7 & 90.6 & 50.4 & 80.6 & 69.4 & NA \\
plain-all 		& 75.9 & 90.4 & 53.0 & 82.4 & 65.1 & 41.7 \\
DISCO-reverse 	& 64.5 & 84.5 & 41.2 & 55.5 & 67.0 & 24.9 \\
DISCO-3D-2D 	& 81.5 & 92.0 & 61.0 & 87.6 & 73.1 & NA \\
DISCO 			& \textbf{81.8} & \textbf{93.4} & 59.0 & 87.7 & \textbf{74.3} & \textbf{45.4} \\
\hline
\end{tabular}
\caption{\small PCK[$\alpha=0.1$] accuracies (\%) of different methods for 2D keypoint localization on the car category of PASCAL VOC. Bold numbers indicate the best results.}
\label{tab:voc}
\end{table*}

We observe that DISCO outperforms competitors in both 2D and 3D keypoint localization across all occlusion types. 
Moreover, we observe a monotonic increase in 2D and 3D accuracy with increasing supervision: plain-2D or plain-3D $<$ DISCO-3D-2D $<$ DISCO-vis-3D-2D $<$ DISCO.     
Further, plain-all is superior to plain-2d and plain-3d, while DISCO exceeds plain-all by $4.4\%$ on 2D-All and $2.4\%$ on 3D-Full. 
These experiments confirm that joint modeling of 3D shape concepts is better than independent modeling. 
Moreover, alternative supervision orders (DISCO-reverse, DISCO-(3D-vis)) are found to be inferior to the proposed order which captures underlying structure between shape concepts. 
Last, DISCO-VGG performs significantly worse than DISCO by $16.0\%$ on 2D-All and $5.6\%$ on 3D-Full, which validates our removal of local spatial pooling and adopt global average pooling.
In conclusion, the proposed deep supervision architecture coupled with intermediate shape concepts improves the generalization ability of CNN.
As more concepts are introduced in the ``correct'' order, we observe improvement in performance.

We also conduct an ablative study of training data with different occlusion types. 
Table~\ref{tab:data_ablation} demonstrates 2D keypoint localization accuracies over different occlusion categories on KITTI-3D given various combination of training data.
``Occ.'' stands for test examples with multi-object occlusions where the occluder is either another car or a different object such as a pedestrian.
As we can see, DISCO trained on fully visible cars alone achieves much worse performance on truncated and occluded test data than when trained on data with simulated truncation and multi-car occlusion.
We observe that multi-car occlusion data is also helpful in modeling truncation cases, and the network trained by multi-car data obtains the second best result on truncated cars.
The best overall performance is obtained by including all three types of examples (no occlusion, multi-car occlusion, truncation), emphasizing the efficacy of our data generation strategy.

Finally, we evaluate DISCO on detection bounding boxes computed from RCNN\cite{girshick2014rich} with IoU$>0.7$ to the groundtruth of KITTI-3D.
``DISCO-Det'' in the last row of Table~\ref{tab:car3d} shows PCK accuracies of DISCO using detection results.
The 2D/3D keypoint localization accuracies even exceeds the performance of DISCO using groundtruth bounding boxes by $3.3\%$ on 2D-All and $0.2\%$ on 3D-All. 

\subsubsection{PASCAL VOC}
\label{sec:voc}

We evaluate DISCO on the PASCAL VOC 2012 dataset for 2D keypoint localization~\cite{yang2011articulated}. Unlike KITTI-3D where car images are captured on real roads and mostly in low resolution, PASCAL VOC contains car images with larger appearance variations and heavy occlusions. 
In Table \ref{tab:voc}, we compare our results with the state-of-the-art \cite{tulsiani2015viewpoints,long2014convnets} on various sub-classes of the test set: fully visible cars (denoted as ``Full''), occluded cars, high-resolution (average size $420$x$240$) and low-resolution images (average size $55$x$30$). Please refer to \cite{tulsiani2015viewpoints} for details of the test setup. 
Note that these methods~\cite{tulsiani2015viewpoints,long2014convnets} are trained on real images, whereas DISCO training exclusively leverages synthetic training data.

We observe that DISCO outperforms \cite{tulsiani2015viewpoints} by $0.6\%$ and $5.1\%$ on PCK at $\alpha=0.1$ and $\alpha=0.2$, respectively. In addition, DISCO is robust to low-resolution images, improving $6.9\%$ accuracy on low-resolution set compared with \cite{tulsiani2015viewpoints}. 
This is critical in real perception scenarios where distant objects are small in images of street scenes.
However, DISCO is inferior on the occluded car class and high-resolution images, attributable to our use of small images ($64$x$64$) for training and the fact that our occlusion simulation does not capture the complex occlusions created by non-car objects such as walls and trees. 
Finally, we compute APK accuracy at $\alpha=0.1$ for DISCO on the same detection candidates used in \cite{tulsiani2015viewpoints}\footnote{We run the source code \cite{tulsiani2015viewpoints} to obtain the same object candidates.}.
We can see that DISCO outperforms \cite{tulsiani2015viewpoints} by $5.1\%$ on the entire car dataset (Full+Occluded).
This suggests DISCO is more robust to noisy detection results and more accurate on keypoint visibility inference than \cite{tulsiani2015viewpoints}.
We attribute this to global structure modeling of DISCO during training where the full set of 2D keypoints resolves the partial view ambiguity whereas traditional methods like \cite{tulsiani2015viewpoints} only are supervised with visible 2D keypoints. 

Note that some definitions of our car keypoints~\cite{zia2013detailed} are slightly different from \cite{yang2011articulated}.
For example, we annotate the bottom corners of the front windshield whereas \cite{yang2011articulated} labels the side mirrors. 
In our experiments, we ignore this annotation inconsistency and directly assess the prediction results.
We reemphasize that unlike \cite{long2014convnets,tulsiani2015viewpoints}, we do not use the PASCAL VOC train set. 
Thus, even better performance is expected when real images with consistent labels are used for training.

\subsubsection{PASCAL 3D+}
\label{sec:pascal3d}

\begin{table}[t]
\small
\centering
\begin{tabular}{| c | c | c |}
\hline
Method 									& CAD alignment GT 	& Manual GT \\
\hline
VDPM-16 \cite{xiang2014beyond}    		& NA   				& 51.9  \\
Xiang et al. \cite{mottaghi2015coarse} 	& 64.4 				& 64.3  \\
Random CAD \cite{xiang2014beyond}       & NA   				& 61.8  \\
GT CAD \cite{xiang2014beyond}   	 	& NA   				& 67.3  \\
\hhline{|=|=|=|}
DSN-2D									& 66.4				& 63.3\\
plain-2D								& 67.4				& 64.3\\
plain-all								& 66.8 				& 64.2\\
DISCO-reverse							& 54.2				& 56.0\\
DISCO    								& \textbf{71.2} 	& \textbf{67.6}  \\
\hline
\end{tabular}
\caption{\small Object segmentation accuracies (\%) of different methods on PASCAL3D+. Best results are shown in bold.}
\label{tab:pascal3d}
\end{table}

PASCAL3D+ \cite{xiang2014beyond} provides object viewpoint annotations for PASCAL VOC objects by manually aligning 3D object CAD models onto the visible 2D keypoints. 
Because only a few CAD models are
 used for each category, the 3D keypoint locations are only approximate. 
 Thus, we use the evaluation metric proposed by \cite{xiang2014beyond} which measures 2D overlap (IoU) against projected model mask. With a 3D skeleton of an object, we are able to create a coarse object mesh based on the geometry and compute segmentation masks by 
 projecting coarse mesh surfaces onto the 2D image based on the estimated 2D keypoint locations.

Table \ref{tab:pascal3d} reports object segmentation accuracies on two types of ground truth. The column ``Manual GT'' uses manual pixel-level 
annotation provided by PASCAL VOC 2012, whereas ``CAD alignment GT'' uses 2D projections of aligned CAD models as ground truth. Note that ``CAD alignment GT'' 
covers the entire object extent in the image including regions occluded by other objects. DISCO significantly outperforms a state-of-the-art method \cite{xiang2015data} 
by $4.6\%$ and $6.6\%$ despite using only synthetic data for training. Moreover, on ``Manual GT'' benchmark, we compare DISCO with ``Random CAD'' and ``GT CAD'' which stand for 
the projected segmentation of randomly selected and ground truth CAD models respectively, given ground truth object pose. 
We find that DISCO yields even superior performance to ``GT CAD''.
This provides evidence that joint modeling of 3D geometry manifold and viewpoint is better than the pipeline of object retrieval plus alignment.
Finally, we note that a forward pass of DISCO only takes less than 10ms during testing, which is far more efficient compared with sophisticated CAD alignment approaches~\cite{zia2013detailed} that usually needs more than 1s for one image input.


\subsubsection{IKEA}
\label{sec:ikea}
\begin{figure*}[t]
  \centering
\begin{subfigure}{0.33\textwidth}
  \centering
  \includegraphics[width=1.0\linewidth]{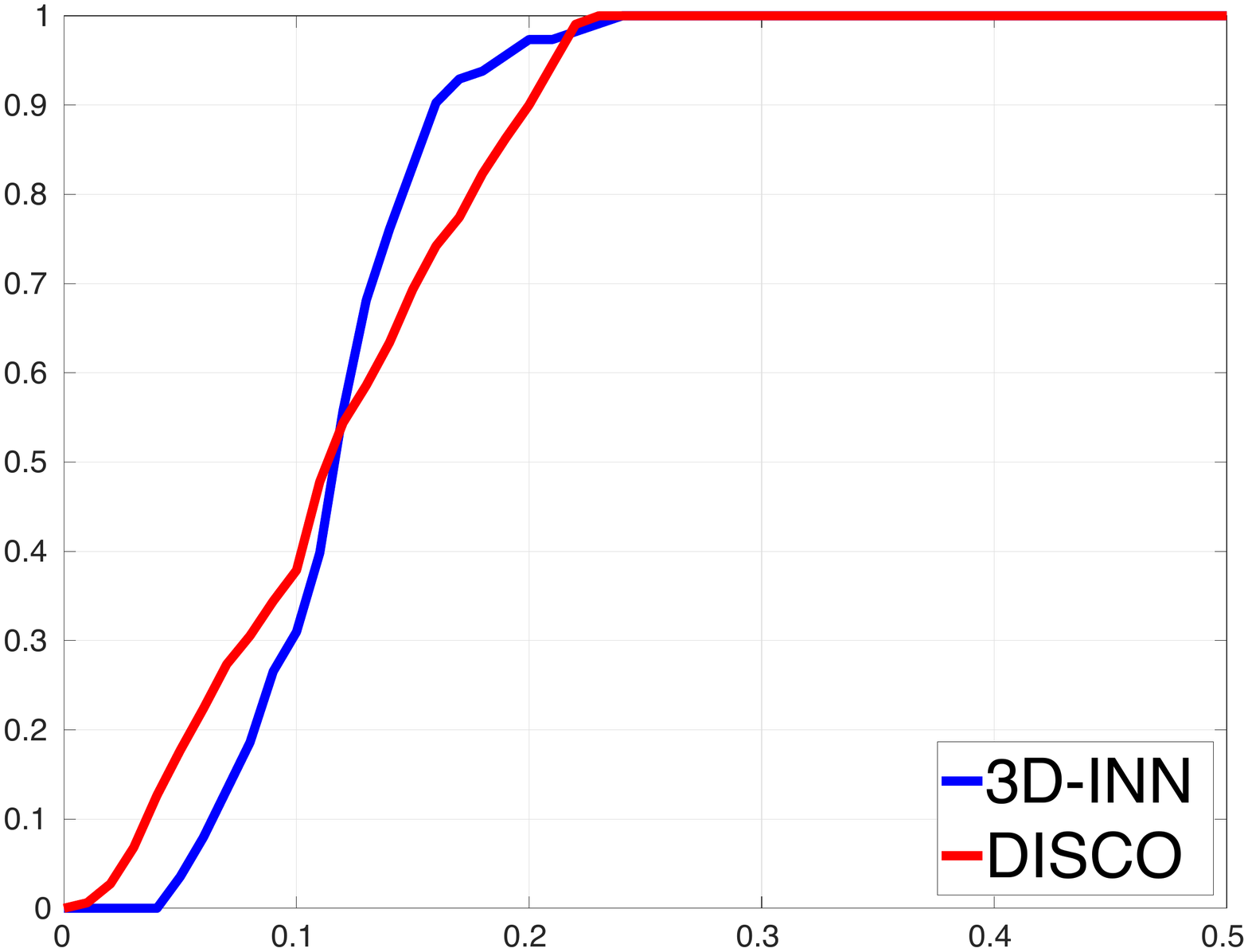}
  \caption{IKEA Sofa}
  \label{fig:pck_sofa_3d}  
\end{subfigure}%
\begin{subfigure}{0.33\textwidth}
  \centering
  \includegraphics[width=1.0\linewidth]{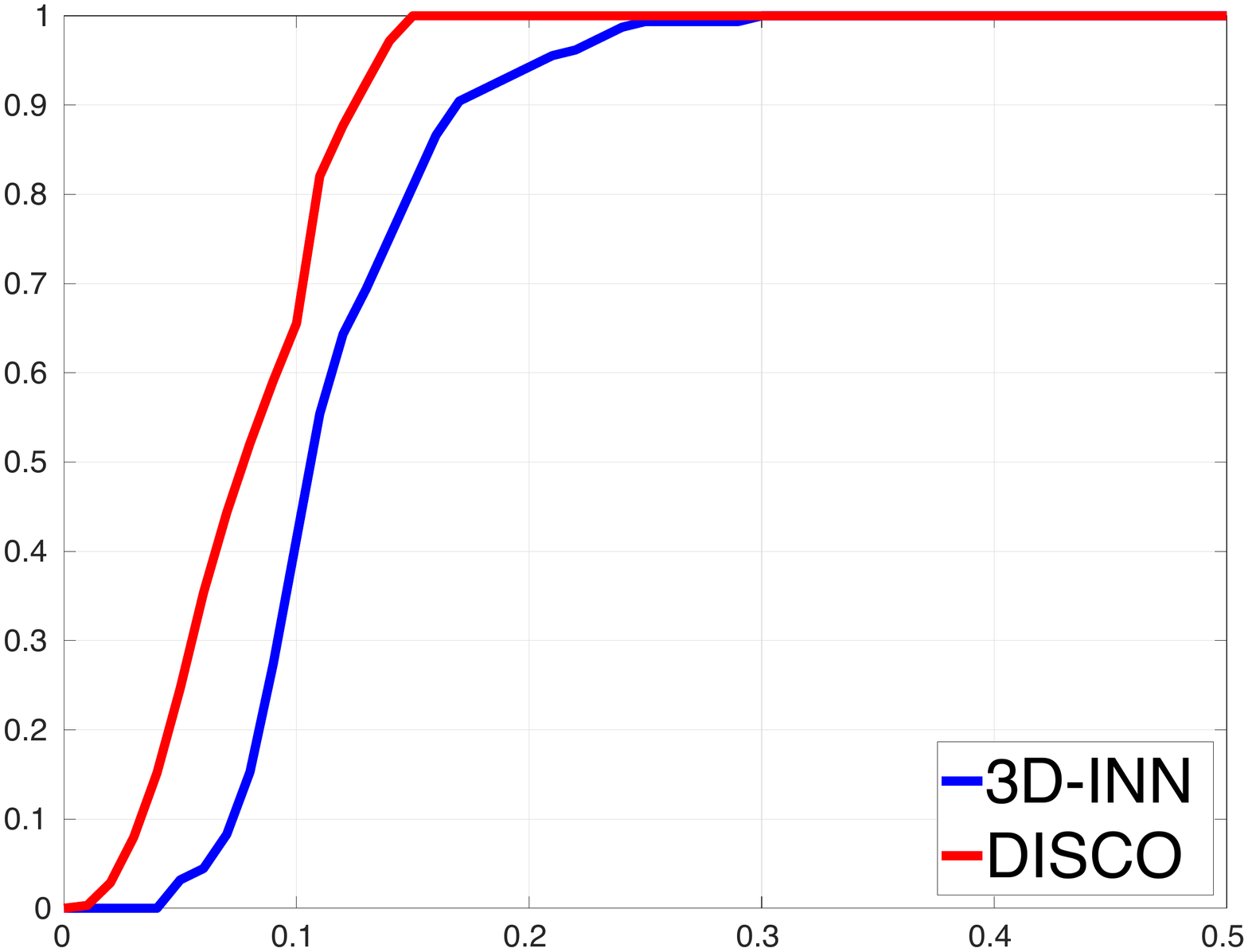}
  \caption{IKEA Chair}
  \label{fig:pck_chair_3d}
\end{subfigure}
\begin{subfigure}{0.33\textwidth}
  \centering
  \includegraphics[width=1.0\linewidth]{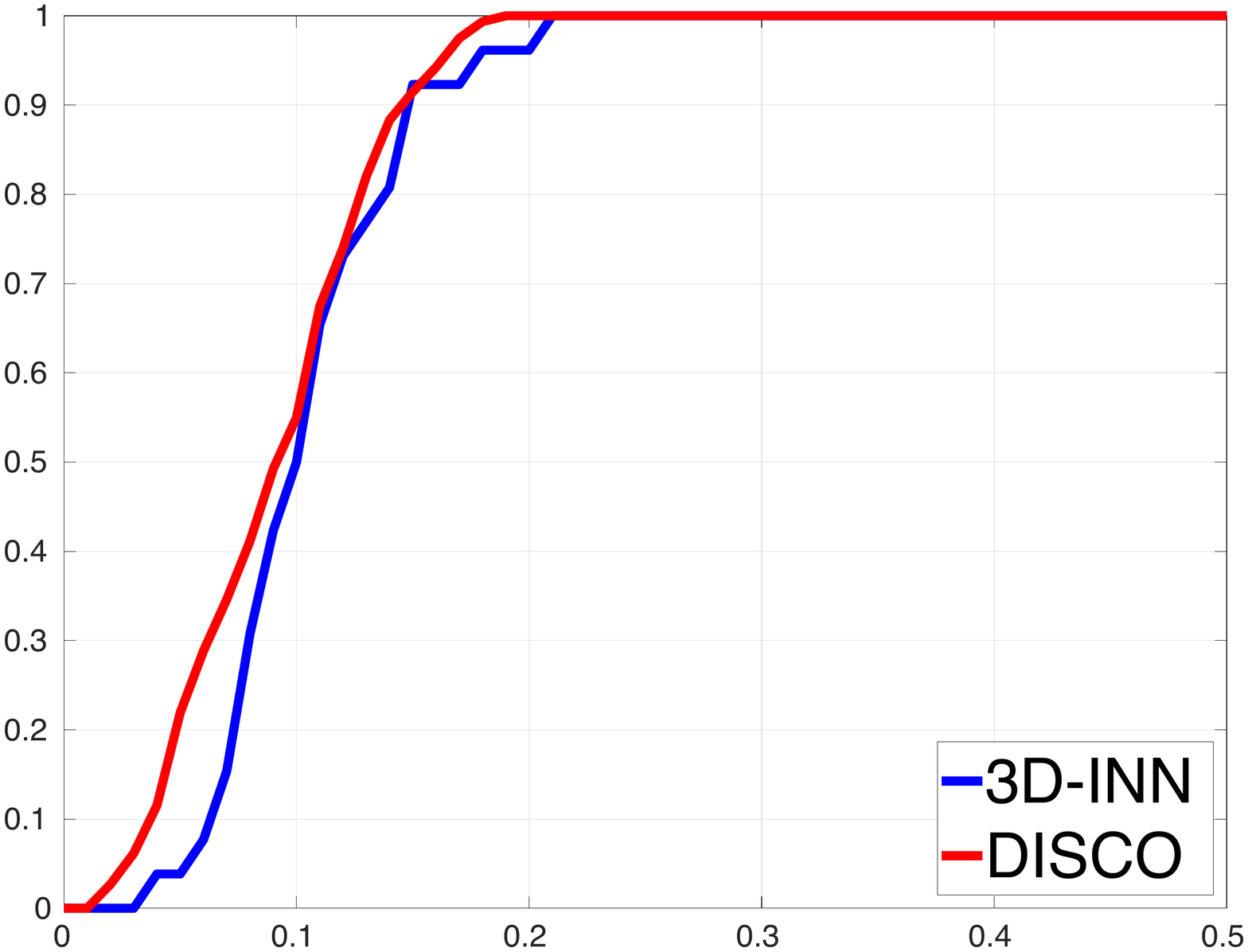}
  \caption{IKEA Bed}
  \label{fig:pck_bed_3d}
\end{subfigure}%
\caption{3D PCK (RMSE\cite{wu2016single}) curves of DISCO and 3D-INN on sofa (Figure \ref{fig:pck_sofa_3d}), chair (Figure \ref{fig:pck_chair_3d}) and bed (Figure \ref{fig:pck_bed_3d}) classes of IKEA dataset. In each figure, X axis stands for $\alpha$ of PCK and Y axis represents the accuracy.
}
\label{fig:pck_ikea}
\end{figure*}

\begin{table}[t]
\small
\centering
\begin{tabular}{| c | c | c || c | c || c | c |}
\hline
\multirow{2}{*}{Method}	&	\multicolumn{2}{c||}{Sofa} & \multicolumn{2}{c||}{Chair} & \multicolumn{2}{c|}{Bed} \\
\hhline{|~|-|-||-|-||-|-|} &  Recall & PCK & Recall & PCK & Recall & PCK  \\
\hline
3D-INN  & \textbf{88.0} & 31.0 & 87.8 & 41.4 & \textbf{88.6} & 42.3 \\
DISCO  & 84.4 & \textbf{37.9} & \textbf{90.0} & \textbf{65.5} & 87.1 & \textbf{55.0} \\
\hline
\end{tabular}
\caption{\small Average recall and PCK[$\alpha=0.1$] accuracy(\%) for 3D structure prediction on the sofa and chair classes on IKEA dataset.
}
\label{tab:ikea}
\end{table}

In this section, we evaluate DISCO on the IKEA dataset \cite{lim2013parsing} with 3D keypoint annotations provided by \cite{wu2016single}.
One question remaining for the DISCO network is whether it is capable of learning 3D object geometry for multiple object classes simultaneously. 
Therefore, we train a single DISCO network from scratch which jointly models three furniture classes: sofa, chair and bed. 
At test time, we compare DISCO with the state-of-the-art 3D-INN\cite{wu2016single} on IKEA.
Since 3D-INN evaluates the error of 3D structure prediction in the object canonical pose, we align the PCA bases of both the estimated 3D keypoints and their groundtruth.
Table \ref{tab:ikea} reports the PCK[$\alpha=0.1$] and average recall\cite{wu2016single} (mean PCK over densely sampled $\alpha$ within $[0,1]$) of 3D-INN and DISCO on all furniture classes.
The corresponding PCK curves are visualized in Figure~\ref{fig:pck_ikea}.
We retrieve PCK accuracies of 3D-INN on the IKEA dataset from its publicly released results.
DISCO significantly outperforms 3D-INN on PCK by $6.6\%,24.1\%,12.7\%$ on sofa, chair and bed respectively, which means that DISCO obtains more correct predictions of keypoint locations than 3D-INN.
This substantiates that direct exploitation of the rich visual details from images adopted by DISCO is critical to infer more accurate and fine-grained 3D structure than lifting sparse 2D keypoints to 3D shapes like 3D-INN. 
However, DISCO is inferior to 3D-INN in terms of average recall on the sofa and bed class.
As shown in Figure~\ref{fig:pck_sofa_3d}, the incorrect predictions by DISCO deviate more from the groundtruth than 3D-INN.
This is mainly because 3D predicted shapes from 3D-INN are constrained by shape bases so even incorrect estimates have realistic object shapes when recognition fails. 
Moreover, our 3D keypoint labeling for the sofa CAD models is slightly different from \cite{wu2016single}.
We annotate the corners of reachable seating areas of a sofa while IKEA labels the corners of the outer volume parallel to the seating area
We conclude that DISCO is able to learn 3D patterns of object classes other than the car category and shows potential as a general-purpose approach to jointly model 3D geometric structure of multiple objects \emph{in a single model}.
 
\subsubsection{Qualitative Results}
\label{sec:quality}

\begin{figure*}[t]
  \centering
  \includegraphics[width=1.0\linewidth]{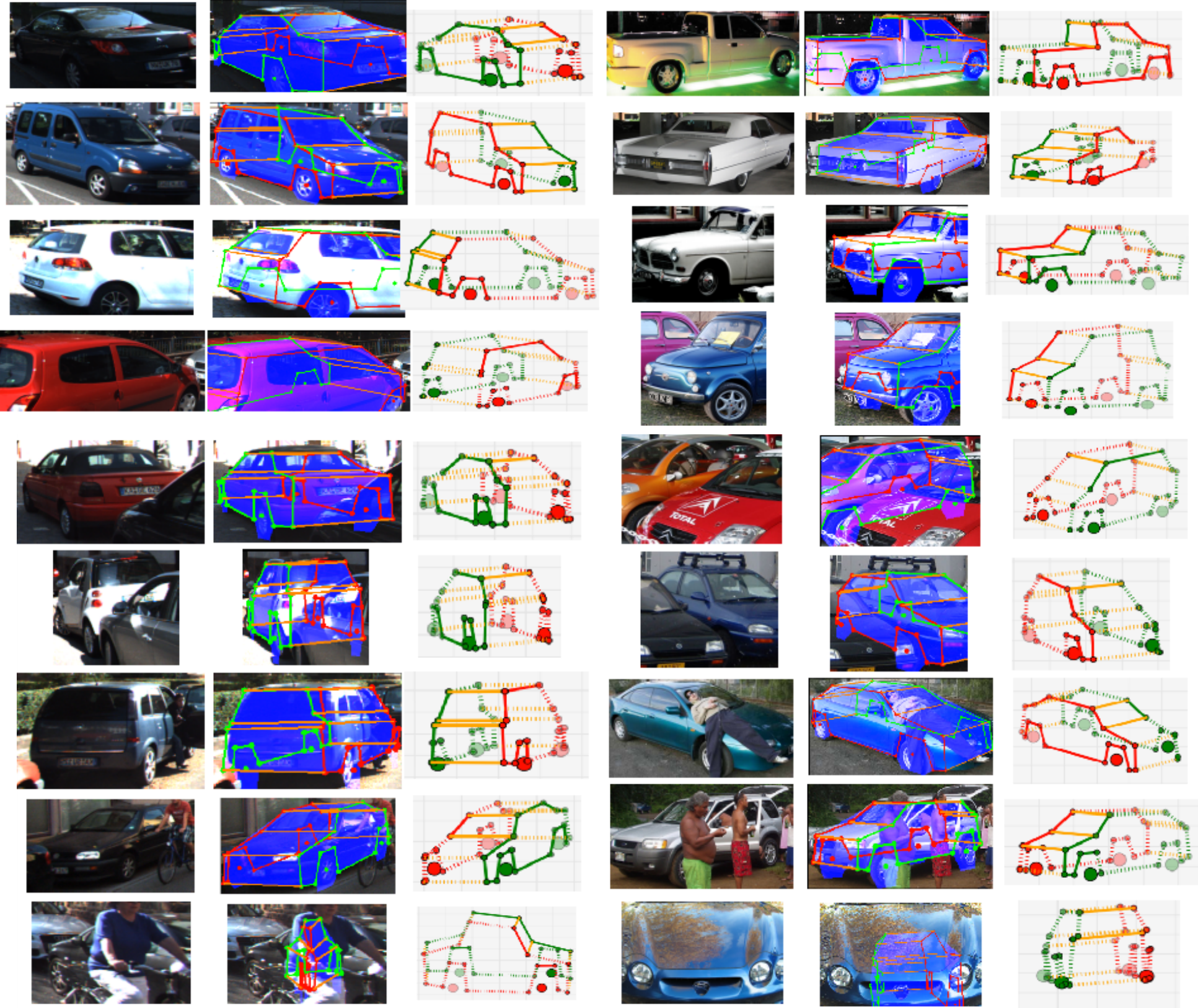}
  \caption{Visualization of 2D/3D prediction, visibility inference and instance segmentation on KITTI-3D (left column) and PASCAL VOC (right column). Last row shows failure cases. Circles and lines represent keypoints and their connections. Red and green indicate the left and right sides of a car, orange lines connect two sides. Dashed lines connect keypoints if one of them is inferred to be occluded. Light blue masks present segmentation results.}
 \label{fig:demo}
\end{figure*}
\begin{figure*}[t]
  \centering
    \includegraphics[width=0.9\linewidth]{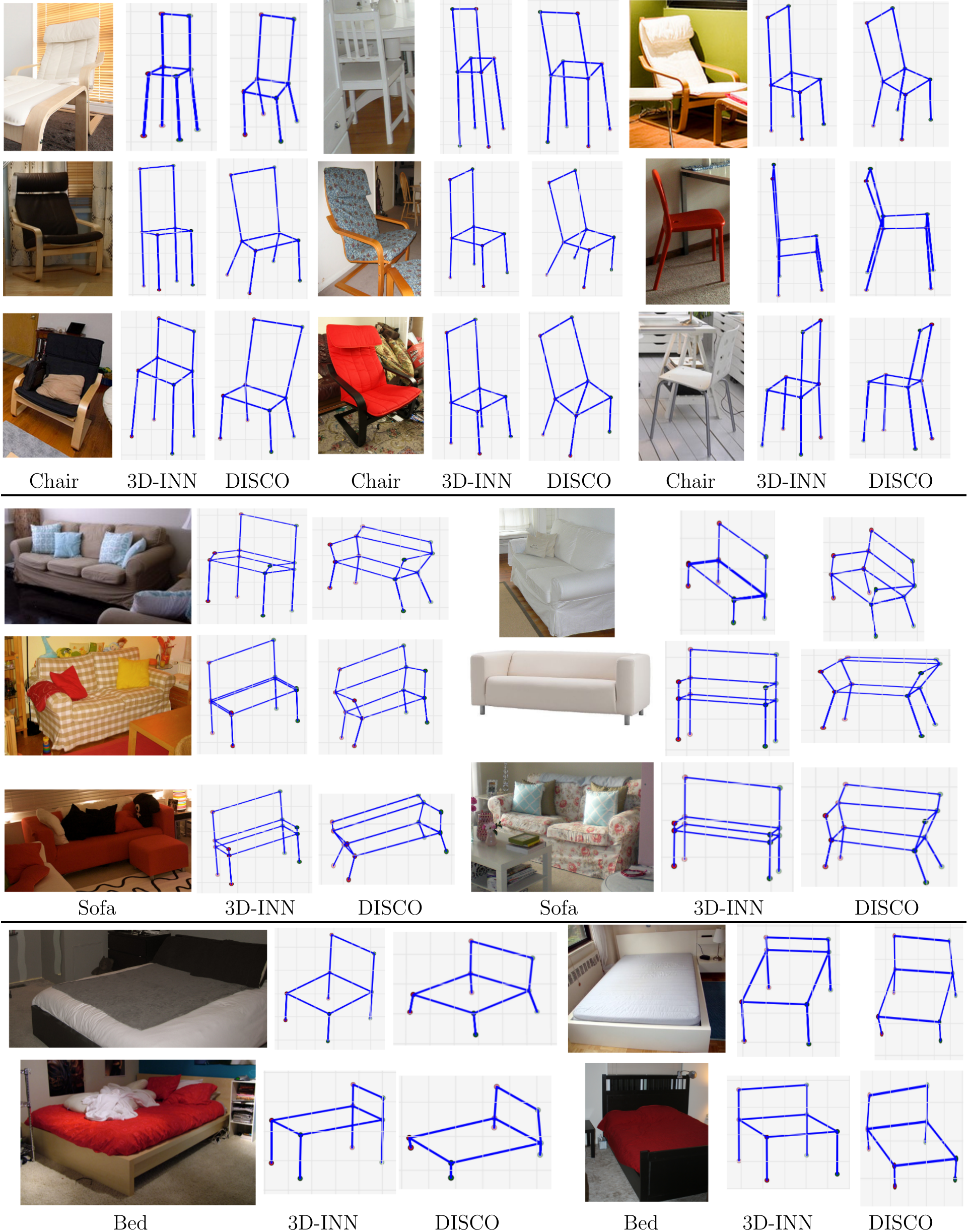}
  \caption{Qualitative comparison between 3D-INN and DISCO for 3D stricture prediction on IKEA dataset.}
 \label{fig:ikea_demo}
\end{figure*}

In Figure \ref{fig:demo}, we visualize example predictions from DISCO on KITTI-3D (left column) and PASCAL VOC (right column). From left to right, each column shows the 
original object image, the predicted 2D object skeleton with instance segmentation and the predicted 3D object skeleton with visibility. 
From top to bottom, we show the results under no occlusion (row 1-2), truncation (row 3-4), multi-car occlusion (row 5-6), other occluders (row 7-8) and failure cases (row 9). 
We observe that DISCO is able to localize 2D and 3D keypoints on real images with complex occlusion scenarios and diverse car models such as sedan, SUV and pickup. 
Moreover, the visibility inference is mostly correct.
These capabilities highlight the potential of DISCO as a building block for holistic scene understanding in cluttered scenes. 
In failure cases, the left car is mostly occluded by another object and the right one is severely truncated and distorted in projection. 
We may improve the performance of DISCO on these challenging cases by exploiting more sophisticated data simulated with complex occlusions~\cite{richter2016playing} and finetuning DISCO on real data.

In addition, we qualitatively compare 3D-INN and DISCO on three categories in IKEA dataset  in Figure \ref{fig:ikea_demo}.
For the chair, 3D-INN fails to delineate the inclined seatbacks in the example images while DISCO being able to capture this structural nuance.
For the sofa, DISCO correctly infers the location of sofa armrest whereas 3D-INN merges armrests to the seating area or predicts an incorrect size of the seatback. 
Finally, DISCO yields better estimates of the scale of bed legs than 3D-INN.
We attribute this relative success of DISCO to direct mapping from image evidence to 3D structure, as opposed to lifting 2D keypoint predictions to 3D.

\section{Conclusion}
\label{sec:con}

Visual perception often involves sequential inference over a series of intermediate goals of growing complexity towards the final objective. 
In this paper, we have employed a probabilistic framework to formalize the notion of intermediate concepts which points to better generalization through deep supervision, compared to the standard end-to-end training. 
This inspires a CNN architecture where hidden layers are supervised with an intuitive sequence of intermediate concepts, in order to incrementally regularize the learning to follow the prescribed inference sequence. 
We practically leveraged this superior generalization capability to address the scarcity of 3D annotation: learning shape patterns from synthetic training images with complex multiple object configurations. 
Our experiments demonstrate that our approach outperforms current state-of-the-art methods on 2D and 3D landmark prediction on public datasets, even with occlusion and truncation. 
We applied deep supervision to fine-grained image classification and showed significant improvement over single-task as well as multi-task networks on CIFAR100. 
Finally, we have presented preliminary results on jointly learning 3D geometry of multiple object classes within a single CNN.
Our future work will extend this direction by learning shared representations for diverse object classes. 
We also see wide applicability of deep supervision, even beyond computer vision, in domains such as robotic planning, scene physics inference and generally wherever deep neural networks are being applied. 
Another future direction is to extract label relationship graphs from the CNN supervised with intermediate concepts, as opposed to explicitly constructed Hierarchy and Exclusion graphs~\cite{Deng2014eccv}.
\section*{Acknowledgments}
This work was part of C. Li's intern project at NEC Labs America, in Cupertino. We acknowledge the support by NSF under Grant No. 1227277. We also thank Rene Vidal, Alan Yuille, Austin Reiter and Chong You for helpful discussions.

\bibliographystyle{IEEEtran}
\bibliography{ref_short}

\vfill

\end{document}